\documentclass[letterpaper]{article} %
\PassOptionsToPackage{dvipsnames}{xcolor}
\usepackage{aaai2026}  %
\usepackage{times}  %
\usepackage{helvet}  %
\usepackage{courier}  %
\usepackage[hyphens]{url}  %
\usepackage{graphicx} %
\usepackage{natbib}  %
\usepackage{caption} %
\usepackage{algorithm}

\usepackage{newfloat}
\usepackage{listings}
\DeclareCaptionStyle{ruled}{labelfont=normalfont,labelsep=colon,strut=off} %
\floatstyle{ruled}
\newfloat{listing}{tb}{lst}{}
\floatname{listing}{Listing}
\nocopyright %

\title{LLM-Evolved Pattern Generators for Optimal Classical Planning}

\author{
    Windy Phung,
    Dominik Drexler,
    Arnaud Lequen,
    Jendrik Seipp
}
\affiliations{
    Link{\"o}ping University, Sweden\\
    firstname.lastname@liu.se
}

\usepackage{algorithm}
\usepackage[noend]{algpseudocode}
\usepackage{booktabs}
\usepackage{multirow,makecell}
\usepackage{amsmath,amssymb,amsthm}
\usepackage{multicol}
\usepackage{stmaryrd}
\usepackage{mathtools}
\usepackage{subcaption}
\usepackage{tikz}
\usetikzlibrary{shapes,arrows,arrows.meta}
\usetikzlibrary{automata}
\usetikzlibrary{positioning}
\usetikzlibrary{backgrounds}
\usetikzlibrary{patterns,patterns.meta,angles}
\usetikzlibrary{graphs, quotes}
\usetikzlibrary{bbox}
\usepackage{forest}
\usepackage{pgfplots}
\pgfplotsset{compat=1.16}
\usepackage[mode=buildnew]{standalone}
\usepackage{paralist}
\usepackage{rotating}

\usepackage[main=american]{babel}
\useshorthands*{"}
\defineshorthand{"-}{\babelhyphen{hard}}
\defineshorthand{"=}{\babelhyphen{–}}
\defineshorthand{"/}{\babelhyphen{/}}

\newcommand{\Omit}[1]{}

\newcommand{\tup}[1]{\langle #1\rangle}

\newcommand{\reals}{\ensuremath{\mathbb{R}}}
\newcommand{\extreals}{\ensuremath{\overline{\mathbb{R}}}}

\newcommand{\defined}[1]{\emph{#1}}

\definecolor{tblue}{RGB}{0,119,187}
\definecolor{tcyan}{RGB}{51,187,238}
\definecolor{tteal}{RGB}{0,153,136}
\definecolor{torange}{RGB}{238,119,51}
\definecolor{tred}{RGB}{204,51,17}
\definecolor{tmagenta}{RGB}{238,51,119}
\definecolor{tgray}{RGB}{187,187,187}

\definecolor{darkgreen}{rgb}{0.0, 0.2, 0.13}
\usepackage{todonotes}
\presetkeys{todonotes}{inline}{}

\newcommand{\eg}{e.g.}

\newcommand{\commentcite}[2]{{%
			\let\leftcite\relax%
			\let\rightcite\relax%
			(#1, \cite{#2})}}
\newcommand{\egcite}[1]{\citep[\eg,][]{#1}}
\newcommand{\inlinecite}[1]{\citeauthor{#1}~\shortcite{#1}}
\newcommand{\name}[1]{\emph{#1}}

\newcommand{\astar}{\ensuremath{\text{A}^*}}

\DeclareMathOperator*{\argmax}{arg\,max}

\newcommand{\task}{\ensuremath{\Pi}}
\newcommand{\vars}{\ensuremath{V}}
\newcommand{\var}{\ensuremath{v}}
\newcommand{\vardomain}{\ensuremath{D}}
\newcommand{\actions}{\ensuremath{A}}
\newcommand{\action}{\ensuremath{a}}
\newcommand{\cost}{\ensuremath{\textit{cost}}}
\newcommand{\costfunctions}{\ensuremath{\mathcal{C}}}
\newcommand{\initial}{\ensuremath{I}}
\newcommand{\goal}{\ensuremath{\gamma}}
\newcommand{\pre}{\ensuremath{\mathit{pre}}}
\newcommand{\eff}{\ensuremath{\mathit{eff}}}

\newcommand{\tsys}{\ensuremath{\mathcal{T}}}
\newcommand{\states}{\ensuremath{S}}
\newcommand{\state}{\ensuremath{s}}
\newcommand{\apply}[1]{\mathopen{\llbracket}#1\mathclose{\rrbracket}}
\newcommand{\tlabels}{\ensuremath{L}}
\newcommand{\tlabel}{\ensuremath{\ell}}
\newcommand{\transitions}{\ensuremath{T}}
\newcommand{\initialstate}{\ensuremath{\state_0}}
\newcommand{\goalstates}{\ensuremath{S_*}}
\newcommand{\plan}{\ensuremath{\pi}}

\newcommand{\heuristics}{\ensuremath{\mathcal{H}}}
\newcommand{\heuristic}{\ensuremath{h}}

\newcommand{\abstraction}{\ensuremath{\alpha}}

\newcommand{\pattern}{\ensuremath{P}}

\newcommand{\sys}[1]{\ensuremath{\textsc{Sys-#1}}}
\newcommand{\sysscp}{\ensuremath{\textsc{Sys-SCP}}}

\newcommand{\hevolved}{\ensuremath{\heuristic^{\text{evo}}}}
\newcommand{\hsys}[1]{\ensuremath{\heuristic^{\textsc{Sys}}_{#1}}}
\newcommand{\hsysone}{\hsys{1}}
\newcommand{\hsystwo}{\hsys{2}}
\newcommand{\hsysthree}{\hsys{3}}
\newcommand{\hsysscp}{\hsys{\text{SCP}}}
\newcommand{\hrand}{\ensuremath{\heuristic^{\text{rand}}}}

\newcommand{\scp}{\ensuremath{\text{SCP}}}
\newcommand{\saturate}{\ensuremath{\mathit{saturate}}}
\newcommand{\rem}{\ensuremath{\textit{rem}}}
\newcommand{\mscf}{\ensuremath{\textit{mscf}}}

\newcommand{\numtasks}[1]{\small{(#1)}}

\definecolor{backcolour}{rgb}{0.95,0.95,0.92}

\definecolor{tab10blue}{HTML}{1f77b4}
\definecolor{tab10orange}{HTML}{ff7f0e}
\definecolor{tab10green}{HTML}{2ca02c}
\definecolor{tab10red}{HTML}{d62728}
\definecolor{tab10purple}{HTML}{9467bd}
\definecolor{tab10brown}{HTML}{8c564b}
\definecolor{tab10pink}{HTML}{e377c2}
\definecolor{tab10gray}{HTML}{7f7f7f}

\pgfplotsset{
    cycle list={
            {color=tab10blue, mark=*},
            {color=tab10orange, mark=x},
            {color=tab10green, mark=triangle*},
            {color=tab10red, mark=diamond*},
            {color=tab10purple, mark=pentagon*},
            {color=tab10brown, mark=+},
            {color=tab10pink, mark=square*},
            {color=tab10gray, mark=asterisk},
        }
}

\algrenewcommand\algorithmicindent{1em}%

\begin{document}

\maketitle

\begin{abstract}
Learned heuristics have recently become a competitive alternative to traditional domain-independent heuristics for satisficing planning. Existing approaches, however, focus on improving search guidance rather than guaranteeing admissibility, which makes them unsuitable for optimal classical planning. We present the first method for learning domain-dependent heuristics that are admissible by design and thus preserve the optimality guarantees of \astar{}  search. Instead of learning a direct mapping from states to heuristic values, we learn to construct abstractions that induce admissible heuristics. We use an LLM-driven evolutionary program-synthesis framework to obtain, for each domain, a program that produces a pattern collection for any task in that domain, and we combine the resulting patterns admissibly via saturated cost partitioning. Empirically, the learned programs encode interpretable domain-specific insights, run with negligible overhead at test time and yield heuristics that match the coverage of state-of-the-art domain-independent baselines on several domains while evaluating each state substantially faster.
\end{abstract}

\section{Introduction}

Classical planning aims to find a plan, a sequence of actions whose execution leads from the initial state to the goal, in a deterministic and fully observable environment. We are interested in \emph{optimal} plans, i.e., plans of minimum cost.
\astar{} search with an admissible heuristic is one of the most successful approaches to finding optimal plans \cite{hart-et-al-ieeessc1968}, and its effectiveness depends on the quality of the heuristic. The strongest admissible heuristics today are \emph{domain-independent}: they derive estimates from a task in isolation, without prior knowledge of its domain \cite{helmert-domshlak-icaps2009,seipp-et-al-jair2020}. Tasks from the same domain, however, share regularities that a \emph{domain-dependent} heuristic could exploit. Existing methods do learn such heuristics from solved instances of a domain \cite{toyer-et-al-aaai2018,stahlberg-et-al-icaps2022,chen-et-al-aaai2024}, but they sacrifice admissibility and thus the optimality guarantees of \astar{}. We close this gap by learning domain-dependent heuristics that are admissible by construction.

Pattern-database heuristics (PDBs) are a core ingredient in the strongest admissible heuristics \cite{edelkamp-ecp2001,haslum-et-al-aaai2007,sievers-et-al-socs2012}. Each pattern projects the task onto a small subset of state variables and abstracts away the rest. Goal distances in the resulting abstract state space are admissible estimates for the original task. The strongest admissible PDB heuristics combine the estimates of a \emph{collection} of patterns \cite{pommerening-et-al-ijcai2013,seipp-ijcai2019}. Larger patterns yield more informative estimates, but the number of abstract states they induce grows exponentially with their size, so each pattern must trade information against the cost of computing and querying it.

In this paper, we learn domain-specific \emph{pattern generators} from example tasks. A generator is a Python program that maps a task description to a pattern collection. We synthesize such generators with OpenEvolve \cite{sharma-misc2025}, an LLM-driven evolutionary framework, using planning performance on small training tasks as the optimization signal. At test time, we apply the learned generator directly to new tasks from the same domain, producing pattern collections with negligible overhead.

To combine the resulting PDB heuristics admissibly, we use \emph{saturated cost partitioning} (\scp) \cite{seipp-et-al-jair2020}, one of the strongest methods for this purpose. \scp{} distributes action costs across the patterns in a collection so that the sum of their estimates remains admissible.
Existing pattern generators search the space of patterns separately for each task and cannot reuse the patterns that worked on one task when faced with another from the same domain. We instead learn generators that capture the domain structure observed across example tasks.

We evaluate our learned generators against five baselines on seven domains from the optimal Autoscale benchmark set \cite{torralba-et-al-aaai2021}. The synthesized generators match or exceed the best baseline in four domains, achieve the highest coverage in two and evaluate each state substantially faster than the strongest baselines.
These results show that learning domain-specific pattern generators is a promising direction for improving the performance of optimal planners, and that LLM-driven program synthesis is a powerful tool for this purpose.

\section{Background}
\label{sec:background}

We introduce transition systems, optimal classical
planning, heuristics, projections, saturated cost partitioning and OpenEvolve. Throughout, $\extreals = \reals\cup\{-\infty,\infty\}$ denotes the
extended real numbers.

\paragraph{Transition Systems.}

A \defined{transition system} $\tsys$ consists of a set of states $\states(\tsys)$, a set of labels
$\tlabels(\tsys)$, a set of transitions $\transitions(\tsys)$ of the form
$\tup{\state,\tlabel,\state'}$ with $\state,\state'\in\states(\tsys)$ and
$\tlabel\in\tlabels(\tsys)$, an initial state $\initialstate(\tsys)\in\states(\tsys)$ and a set of
goal states $\goalstates(\tsys)\subseteq\states(\tsys)$. An $\state$-path in a transition system
$\tsys$ is a sequence $\plan = \tup{\state_1,\tlabel_1,\state_2,\ldots,
        \state_n,\tlabel_n,\state_{n+1}}$ with $\state = \state_1$ and
$\tup{\state_i,\tlabel_i,\state_{i+1}}\in\transitions(\tsys)$ for all $i=1,\ldots,n$. It is an
$\state$-goal-path if $\state_{n+1}\in\goalstates(\tsys)$.

A \defined{weighted transition system} is a tuple $\tup{\tsys,\cost}$, where $\tsys$ is a transition
system and $\cost : \tlabels(\tsys)\rightarrow\extreals$ is a cost function. We write $\costfunctions(\tlabels(\tsys))$
for the set of all such cost functions. The cost of a path $\plan =
    \tup{\state_1,\tlabel_1,\state_2,\ldots, \state_n,\tlabel_n,\state_{n+1}}$ is
$\cost(\plan) = \sum_{i=1}^n\cost(\tlabel_i)$. The
goal distance $\heuristic_\tsys^\star(\cost,\state)$ in
$\tup{\tsys,\cost}$ is the minimum cost of any $\state$-goal-path.

\paragraph{Optimal Classical Planning.}

A lifted planning task consists of a domain and a problem instance specified in first-order logic \cite{mcdermott-et-al-tr1998,haslum-et-al-2019}.
The domain defines predicate symbols and action schemas, and the problem instance defines the objects, the initial state and the goal. Instantiating the schemas over the objects yields ground atoms and actions, and standard PDDL normalization and translation \cite{helmert-aij2009} turn the result into a finite-domain ground planning task.
We work with collections of lifted tasks that share a single domain; predicates and action schemas are therefore common to all tasks in a collection, while the initial state and the goal vary.

A (ground) \defined{classical planning task} is a tuple $\task = \tup{\vars, \actions, \allowbreak \cost, \initial,
        \goal}$, defined as follows. $\vars$ is a set of binary variables $\var$, each with domain
$\vardomain_\var = \{\text{true}, \text{false}\}$.
We use binary rather than finite-domain variables so that each variable corresponds to a single propositional fact. Our hypothesis is that this makes it easier for the LLM to learn generators that generalize across tasks.
A \defined{partial variable assignment} is a function
$\sigma : \vars'\rightarrow \bigcup_{\var\in\vars} \vardomain_\var$ with
$\vars'\subseteq\vars$ that satisfies $\sigma(\var)\in\vardomain_{\var}$ for all
$\var\in\vars'$; it is \defined{complete} if $\vars'$ equals $\vars$.
$\actions$ is a set of actions, each of the form $\tup{\pre,\eff}$ where $\pre$ and $\eff$
are partial variable assignments, and $\cost : \actions\rightarrow\reals_0^+$
assigns each action a real-valued, non-negative cost. A \defined{state} is a complete variable assignment;
$\initial$ is the initial state and $\goal$ is a partial variable assignment describing the
goal. An action $\tup{\pre,\eff}$ is \defined{applicable} in a state $\state$ iff
$\state[\var] = \pre(\var)$ for every $\var$ on which $\pre$ is defined. Applying $\action$
to $\state$ then yields the successor state $\state\apply{\action} = \state'$, with $\state'[\var] = \eff(\var)$
if $\eff(\var)$ is defined and $\state'[\var] = \state[\var]$ otherwise.

A task $\task = \tup{\vars, \actions, \cost, \initial, \goal}$ induces the \defined{weighted
    transition system} $\tup{\tsys,\cost}$, where $\states(\tsys)$ is the set of all states over
$\vars$, $\tlabels(\tsys) = \actions$ and $\tup{\state,\action,\state'}\in\transitions(\tsys)$
iff $\state' = \state\apply{\action}$. The initial state is $\initialstate(\tsys) = \initial$, and
$\state\in\goalstates(\tsys)$ iff $\goal(\var) = \state[\var]$ for every $\var$ on which $\goal$
is defined. The objective of optimal classical planning is to find an $\initialstate$-goal-path of
minimum cost in $\tup{\tsys,\cost}$.

\paragraph{Heuristics.}

A \defined{heuristic} is a function $\heuristic_\tsys :
    \costfunctions(\tlabels(\tsys))\times\states(\tsys)\rightarrow\extreals$ that estimates
the cost of a minimum $\state$-goal-path for a state $\state\in\states(\tsys)$ under a
given cost function. A heuristic $\heuristic_\tsys$ is
\defined{admissible} if $\heuristic_\tsys(\cost,\state)\leq\heuristic_\tsys^\star(\cost,\state)$
for all $\cost\in\costfunctions(\tlabels(\tsys))$ and all $\state\in\states(\tsys)$. The
$\state$-goal-path returned by \astar{} search with an admissible heuristic has minimum cost \cite{hart-et-al-ieeessc1968}. We therefore aim to construct accurate
admissible heuristics for optimal classical planning.

\paragraph{Projections.}

Projections that simplify the task are a natural source of admissible heuristics \cite{culberson-schaeffer-cscsi1996,edelkamp-ecp2001}. A \defined{pattern}
$\pattern\subseteq\vars$ is a subset of the task's variables. Let
$\states$ denote the set of states over $\vars$ and $\states'$ the set of states over the variables in
$\pattern$. The pattern induces a \defined{projection} $\abstraction : \states\rightarrow\states'$ with
$\abstraction(\state) = \state'$ iff $\state[\var] = \state'[\var]$ for all $\var\in\pattern$. This
projection in turn induces an abstract transition system $\tsys'$ of the transition system $\tsys$
of $\task$, defined by (1) $\abstraction(\initialstate(\tsys)) = \initialstate(\tsys')$, (2)
$\tup{\abstraction(\state),\tlabel,\abstraction(\state')}\in\transitions(\tsys')$ whenever
$\tup{\state,\tlabel,\state'}\in\transitions(\tsys)$ and (3)
$\abstraction(\state)\in\goalstates(\tsys')$ whenever $\state\in\goalstates(\tsys)$. Setting
$\heuristic_{\tsys}(\cost,\state)$ to the minimum cost of an $\abstraction(\state)$-goal-path in
$\tup{\tsys',\cost}$ yields an admissible heuristic. This heuristic is uninformative when the pattern is too small and intractable to compute when it is too large. The standard remedy is to construct a \emph{collection} of patterns and combine their estimates admissibly with cost partitioning.

\paragraph{Saturated Cost Partitioning.}

A \emph{cost partitioning} \cite{katz-domshlak-aij2010} of a cost function $\cost\in\costfunctions(\tlabels(\tsys))$
is a tuple $\tup{\cost_1,\ldots,\cost_n}\in\costfunctions(\tlabels(\tsys))^n$ such that $\sum_{i=1}^n
    \cost_i(\tlabel)\leq \cost(\tlabel)$ for all $\tlabel\in\tlabels(\tsys)$. Given a collection of abstraction heuristics $\heuristics =
    \tup{\heuristic_1,\ldots,\heuristic_n}$, the corresponding \defined{cost partitioning heuristic} is
$\heuristic^{\text{CP}}(\cost,\state) = \sum_{i=1}^n \heuristic_{\tsys_i}^\star(\cost_i,\state)$, where $\tsys_i$
is the abstract transition system underlying $\heuristic_i$.

\defined{Saturated cost partitioning} (SCP) \cite{seipp-et-al-jair2020} is a greedy method for computing such cost partitionings.
Given a sequence of heuristics $\heuristics = \tup{\heuristic_1,\ldots,\heuristic_n}$, it constructs a cost
partitioning $\tup{\cost_1,\ldots,\cost_n}\in\costfunctions(\tlabels(\tsys))^n$ via the recurrence
\begin{align*}
    \rem_0  & = \cost                              &  &                                 \\
    \cost_i & = \saturate(\heuristic_i,\rem_{i-1}) &  & \text{for all } 1\leq i\leq n   \\
    \rem_i  & = \rem_{i-1} - \cost_i               &  & \text{for all } 1\leq i \leq n,
\end{align*}
where
infinite remaining costs are sticky ($\rem_{i-1}(\tlabel) = \infty$ implies $\rem_i(\tlabel) = \infty$), and
$\saturate$ returns the \emph{minimum saturated cost function} $\mscf_i$ for heuristic
$\heuristic_i$ with abstract transition system $\tsys_i$ under remaining costs
$c$:
\begin{multline*}
    \mscf_i(\tlabel) = \sup_{\substack{\tup{\state,\tlabel,\state'}\in\transitions(\tsys_i) \\
            \text{s.t.\ }\heuristic_{\tsys_i}^\star(c,\state)<\infty}}
    \bigl(\heuristic_{\tsys_i}^\star(c,\state) - \heuristic_{\tsys_i}^\star(c,\state')\bigr)
\end{multline*}

Standard enhancements include using multiple diverse orders \cite{seipp-et-al-aaai2017} and saturating over only a subset of states \cite{seipp-helmert-icaps2019}. The exact details are not crucial here because we use the same state-of-the-art configuration of \scp{} throughout.

\paragraph{OpenEvolve.}
OpenEvolve \cite{sharma-misc2025} is an open-source framework for evolving programs with large language models (LLMs), modeled after DeepMind's AlphaEvolve \cite{novikov-et-al-arxiv2025}.
A \defined{program} $p$ is executed to produce outputs; a quality function $q(p)\in\reals$
scores those outputs, and a feature extractor returns a $d$-dimensional vector
$\phi(p)\in\reals^d$ characterizing the program's behavior.

\begin{figure}[t]
    \centering
    \begin{tikzpicture}[
        >={Stealth[length=4pt]},
        font=\footnotesize,
        cell/.style={draw=gray!50, line width=0.3pt, minimum size=2.4mm, inner sep=0pt},
        island/.style={draw=black!60, rounded corners=2pt, fill=blue!4,
                inner sep=2pt, line width=0.5pt},
        migrate/.style={->, dashed, draw=gray!60!black, line width=0.5pt},
        flow/.style={->, draw=black!75, line width=0.5pt},
        box/.style={draw=black!60, rounded corners=2pt, fill=orange!8,
                inner sep=2pt, line width=0.5pt, align=center, minimum height=4mm}
        ]
        \foreach \name/\posx/\posy/\filled in {
        I1/0/1.7/{0/2,1/1,2/2,3/1,2/3},
        I2/-1.9/-0.4/{0/1,1/3,2/0,3/2},
        I3/1.9/-0.4/{1/0,2/2,0/3,3/1,1/2}}
        {
        \begin{scope}[shift={(\posx,\posy)}]
            \node[island, minimum width=12mm, minimum height=12mm] (\name) at (0,0) {};
            \foreach \x in {0,1,2,3}
            \foreach \y in {0,1,2,3}
                {\node[cell] at ({-3.6mm + 2.4mm*\x}, {-3.6mm + 2.4mm*\y}) {};}
            \foreach \fx/\fy in \filled {
                \node[cell, fill=teal!55] at
                ({-3.6mm + 2.4mm*\fx}, {-3.6mm + 2.4mm*\fy}) {};
            }
        \end{scope}
        }
        \node[anchor=south] at (I1.north) {$I_1$};
        \node[anchor=north] at (I2.south) {$I_2$};
        \node[anchor=north] at (I3.south) {$I_3$};
        \draw[migrate] (I1) to[bend right=12] (I2);
        \draw[migrate] (I2) to[bend right=12] (I1);
        \draw[migrate] (I2) to[bend right=12] (I3);
        \draw[migrate] (I3) to[bend right=12] (I2);
        \draw[migrate] (I3) to[bend right=12] (I1);
        \draw[migrate] (I1) to[bend right=12] (I3);
        \node[gray!60!black] at (0,0.05) {\scriptsize migrate every $\Delta$};
        \node[box, anchor=west] (parent) at (3, 1.8) {parent $p_{\mathrm{par}}$};
        \node[box, anchor=west, fill=blue!8] (llm) at (3.0, 0.9) {LLM};
        \node[box, anchor=west] (off) at (3.0, 0.0) {offspring $p'$};
        \node[box, anchor=west] (place) at (3.0, -0.9) {place in $G_j[c(p')]$ \\ if cell empty or better};
        \draw[flow] (parent) -- (llm)
        node[midway, right, font=\scriptsize] {\,+ inspirations};
        \draw[flow] (llm) -- (off);
        \draw[flow] (off) -- (place);
        \draw[flow, dotted] (I3.east) to[out=0, in=180] (parent.west);
    \end{tikzpicture}
    \caption{One iteration of OpenEvolve. Each island $I_j$ holds a MAP-Elites grid; filled cells contain elite programs. A parent is sampled from the current island, the LLM is prompted with the parent together with the top programs of $I_j$ and \emph{inspirations} drawn from cells near $c(p_{\mathrm{par}})$, and the resulting offspring replaces the occupant of its grid cell if it scores higher. Every $\Delta$ iterations, the top $\rho$-fraction of each island migrates to its two ring neighbors (dashed arrows).}
    \label{fig:openevolve}
\end{figure}
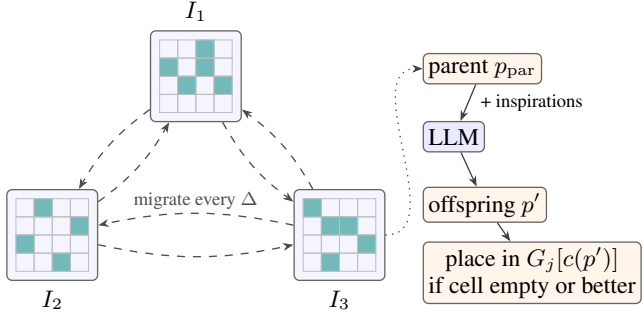

The main loop of OpenEvolve is illustrated on Figure~\ref{fig:openevolve}. At its core, OpenEvolve runs MAP-Elites \cite{mouret-clune-arxiv2015} on a $d$-dimensional grid $G$ of programs. Each program $p$ is assigned to a cell
$c(p) \in \mathbb{Z}^d$ determined by its feature vector $\phi(p)$: the raw features are first
scaled to $[0,1]$ via online min--max normalization, then mapped to discrete bin
indices. A newly generated program $p'$ competes only with the occupant of
$G[c(p')]$ and replaces it whenever the cell is empty or $q(p') > q(G[c(p')])$.

To prevent premature convergence and broaden exploration, OpenEvolve organizes its population into $k$
semi-isolated \defined{islands} $\mathcal{I} = \{I_1, \ldots, I_k\}$, each with its own MAP-Elites
grid $G_j$. The islands form a ring: every $\Delta$ iterations, the top fraction $\rho$ of each island's programs migrate to its two neighbors $G_{(j-1)\bmod k}$ and $G_{(j+1)\bmod k}$, letting good programs spread across islands while still allowing each island to explore a different region of program space.

Within the chosen island $I_j$, a parent $p_{\mathrm{par}}$ is drawn by a mixed strategy: with probability $p_{\mathrm{explore}}$ uniformly at random from the island (exploration), with probability $p_{\mathrm{exploit}}$ from a global archive of elite programs (exploitation) and otherwise by fitness-weighted sampling. The LLM is then prompted with $p_{\mathrm{par}}$'s code together with the top-scoring programs of $I_j$.
It returns an offspring $p'$ that builds on the strengths of these examples while preserving population diversity. Algorithm~\ref{alg:openevolve} summarizes the full procedure.

\begin{algorithm}[t]
    \caption{The OpenEvolve evolutionary loop maintains $k$ islands and selects, mutates and replaces programs across them.}
    \label{alg:openevolve}
    \begin{algorithmic}[1]
        \Require LLM, quality function $q$, feature extractor $\phi$,
        number of islands $k$, migration interval $\Delta$,
        migration fraction $\rho$, iterations $N$
        \Ensure Best program $p^*$
        \State Initialize grids $G_1,\ldots,G_k$;
        seed $G_1[c(p_0)] \leftarrow p_0$
        \For{$i = 1, \ldots, N$}
        \State $j \leftarrow$ island with free capacity
        \State Select parent $p_{\mathrm{par}}$ from $G_j$
        \State Query LLM with $p_{\mathrm{par}}$, $\mathrm{top}(G_j)$
        and inspirations near $c(p_{\mathrm{par}})$ in $G_j$
        to get offspring $p'$
        \State Execute $p'$; compute $c(p')$ and $q(p')$
        \If{$G_j[c(p')] = \emptyset$
        \textbf{ or } $q(p') > q(G_j[c(p')])$}
        \State $G_j[c(p')] \leftarrow p'$
        \EndIf
        \If{$i \bmod \Delta = 0$}
        \ForAll{islands $j'\in\{1,\ldots,k\}$}
        \State Migrate top $\rho$-fraction of $G_{j'}$
        into $G_{(j'-1)\bmod k}$ and $G_{(j'+1)\bmod k}$
        \EndFor
        \EndIf
        \EndFor
        \State \Return $p^* = \argmax_{p \in \bigcup_j G_j} q(p)$
    \end{algorithmic}
\end{algorithm}

\section{Learning to Generate Pattern Collections}
\label{sec:pipeline}
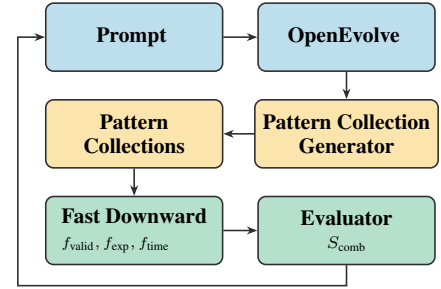
\begin{figure}[tb]
      \centering
      \definecolor{engine}{HTML}{BDDFED}   %
      \definecolor{artifact}{HTML}{FBE5AC} %
      \definecolor{eval}{HTML}{B5E0CB}     %
      \definecolor{border}{HTML}{2F2F2F}   %

      \begin{tikzpicture}[scale=0.9, transform shape, node distance=0.4cm and 0.5cm, font=\footnotesize]
            \node[rectangle, rounded corners=3pt, draw=border, line width=0.6pt, align=center,
                  fill=engine, minimum width=2.6cm, minimum height=1.0cm] (prompt) { \textbf{Prompt}
            };

            \node[rectangle, rounded corners=3pt, draw=border, line width=0.6pt, align=center,
                  fill=artifact, minimum width=2.6cm, minimum height=1.0cm, below=of prompt]
            (patterns) { \textbf{Pattern} \\ \textbf{Collections} };

            \node[rectangle, rounded corners=3pt, draw=border, line width=0.6pt, align=left,
                  fill=eval, minimum width=2.6cm, minimum height=1.0cm, below=of patterns]
            (fastdownward) { \textbf{Fast Downward} \\
                  \tiny $f_{\text{valid}}, f_{\text{exp}}, f_{\text{time}}$ };

            \node[rectangle, rounded corners=3pt, draw=border, line width=0.6pt, align=center,
                  fill=engine, minimum width=2.6cm, minimum height=1.0cm, right=of prompt]
            (OpenEvolve) { \textbf{OpenEvolve} };

            \node[rectangle, rounded corners=3pt, draw=border, line width=0.6pt, align=center,
                  fill=artifact, minimum width=2.6cm, minimum height=1.0cm, below=of OpenEvolve]
            (generator) { \textbf{Pattern Collection} \\ \textbf{Generator} };

            \node[rectangle, rounded corners=3pt, draw=border, line width=0.6pt, align=center,
                  fill=eval, minimum width=2.6cm, minimum height=1.0cm, right=of fastdownward]
            (evaluator) { \textbf{Evaluator} \\ \scriptsize {$S_{\text{comb}}$} };

            \begin{scope}[-{Stealth[length=4pt]}, line width=0.6pt, draw=border]
                  \draw (prompt) -- (OpenEvolve); \draw (generator) -- (patterns); \draw
                  (fastdownward) -- (evaluator);

                  \draw (OpenEvolve.south) -- (generator.north); \draw (patterns.south) --
                  (fastdownward.north);

                  \draw (evaluator.south) -- ++(0,-0.3) -| ([xshift=-0.4cm]fastdownward.west) |-
                  (prompt.west);
            \end{scope}
      \end{tikzpicture}
      \caption{The evolutionary pipeline. The engine (blue) creates artifacts (orange) that are assessed by the evaluation components (green).}
      \label{fig:pipeline}
\end{figure}

For each domain, we learn one \emph{pattern collection generator}: a Python function that maps a task from the domain to a pattern collection for that task. Framing the output as a program rather than a fixed collection lets a single generator handle tasks of any size within its domain at test time. We use the OpenEvolve framework to iteratively improve candidate generators for the domain via LLM-guided mutation, with a scoring function that measures planning performance on a small set of training tasks from the domain. Figure~\ref{fig:pipeline} gives an overview of the resulting pipeline; we describe its components below.

\paragraph{Pattern Collection Generator.}

The pattern collection generator takes a \emph{task information} as input and returns a list of
patterns. The task information provides a structured view of the planning task: static atoms describe its time-invariant properties, fluent atoms describe its dynamic properties, and the initial and goal conditions are given as the atoms that hold in the respective (partial) states.

\paragraph{Scoring.}

To evaluate a generator, we use the combined score $S_{\text{comb}}$ as the quality function
$q$ for Algorithm~\ref{alg:openevolve}. It is the average of
per-problem scores across the set of training tasks $\mathcal{D}$:
\begin{equation*}
      S_{\text{comb}} = \frac{1}{|\mathcal{D}|} \sum_{t \in \mathcal{D}} S_{\text{problem}}(t)
\end{equation*}
The per-problem score combines a validity indicator with normalized efficiency metrics:
\begin{equation}
      S_{\text{problem}} = S_{\text{valid}} \cdot (1 + w_{\text{exp}} \cdot S_{\text{exp}} + w_{\text{time}} \cdot S_{\text{time}}). \nonumber
      \label{eq:scoring_function}
\end{equation}
$S_{\text{valid}}$ is 1 if the generated pattern collection is valid (i.e., it satisfies the constraints from Section~\ref{sec:exp}) and 0 otherwise, so invalid
outputs are penalized immediately. For valid collections, $S_{\text{exp}}$ and $S_{\text{time}}$ measure performance relative to expected bounds on node expansions and search time, with weights $w_{\text{exp}}$ and $w_{\text{time}}$. We use logarithmic scaling:
\begin{equation}
      S_{\text{metric}} = \frac{\log(f_\text{metric}) - \log(\text{upper\_bound})}{\log(\text{lower\_bound}) - \log(\text{upper\_bound})}, \nonumber
\end{equation}
where $f_\text{metric}$ is the observed value and the lower and upper bounds delimit the range of expected
performance. Unsolved tasks contribute $S_{\text{exp}} = S_{\text{time}} = 0$, leaving a base score of 1.

\paragraph{Pipeline.}

We seed OpenEvolve with a single prompt that gives the LLM (i) the domain definition in PDDL form, (ii) the required function signature
\texttt{generate\_pattern\_collection(task\_info: TaskInformation) -> list[Pattern]}, (iii)
task information for three tasks from the domain's training set (the easiest, the median and the hardest by difficulty)
and (iv) a naive baseline implementation that places each goal atom
into its own pattern. Listing~\ref{fig:classes} declares the data structures available to the generator: \texttt{Object}s, \texttt{Predicate}s and \texttt{GroundAtom}s describe the planning task; a \texttt{TaskInformation} bundles the static, initial-state and goal atoms together with all fluent atoms of the task; and a \texttt{Pattern} is a list of ground atoms whose joint projection defines the state space of the corresponding PDB heuristic.

\begin{figure}[!h]
      \input{figures/classes_listing.tex}
      \captionof{lstlisting}{Classes provided in the prompt to the LLM.}
      \label{fig:classes}
\end{figure}

From this prompt, OpenEvolve evolves candidate generators. To evaluate a candidate, we apply it to each training task in order of increasing difficulty and pass the resulting pattern collection to Fast Downward \cite{helmert-jair2006}, which returns the metrics needed to compute the score. Before each run, the collection is validated against size and memory constraints (detailed in Section~\ref{sec:exp}). A task is marked unsolved if the collection violates these constraints, contains a syntactic error or exhausts the time or memory limit, at which point evaluation terminates and the remaining tasks receive a default score. This early termination quickly discards generators that fail on easy instances, focusing evolutionary effort on promising candidates.

\section{Experiments}
\label{sec:exp}
\paragraph{Benchmark Sets.}

We train one generator per domain on a subset of the benchmarks from the Learning track of the International Planning Competition (IPC) 2023 \cite{taitler-et-al-aimag2024}, and we use the Autoscale benchmark set \cite{torralba-et-al-icaps2021} as the test set, which provides instances of increasing size.
We restrict ourselves to the domains that appear in both sets: Blocksworld, Childsnack, Floortile, Miconic, Rovers, Satellite and Transport.
We exclude Sokoban because its PDDL encoding differs between the two sets.

For training, we use every task in the ``easy'' training set and every third task from the ``easy'', ``medium'' and ``hard'' test sets, for a total of 129 tasks per domain.
This is large enough to capture domain diversity, yet small enough to keep the evolutionary loop fast.

\paragraph{OpenEvolve Configuration.}
The evolutionary algorithm runs on 3 islands for 100 iterations using Kimi~K2.5 \cite{kimi-team-arxiv2026} as the underlying LLM.
Each candidate is scored with the formula from Section~\ref{sec:pipeline}, using equal weights $w_{\text{exp}} = w_{\text{time}} = 1$, bounds $[100, 10^6]$ on the number of expansions and bounds $[1, 180]$ on search time in seconds.
Island placement uses a 3-dimensional grid over average expansions before the last $f$-layer, average search time and coverage.

\paragraph{Evaluation.}
We first validate each generated collection: it must contain at most 20 patterns, and each pattern's state space must not exceed $5\cdot 10^6$ states, which keeps PDB computation within a 2\,GiB memory budget.
We then evaluate validated candidates with the Scorpion planner \cite{seipp-ecai2024} under a 3-minute time limit and 4\,GiB of memory per task. All heuristics in our experiments share the same state-of-the-art \scp{} configuration: a diverse set of greedy orders \cite{seipp-et-al-icaps2017} computed online during search \cite{seipp-icaps2021} using the perim* saturator \cite{seipp-helmert-icaps2019}, with a 10-second budget for order generation and a new order computed every 1\,000 expansions.
For each domain, we evaluate the final best generator on 30 test tasks and compare the resulting heuristic, \hevolved{}, against five baselines.

The first four baselines rely on systematic enumeration of \emph{interesting} patterns, i.e., patterns that cannot be replaced by a set of smaller patterns that are equally informative. For $n \in \{1,2,3\}$, \sys{$n$} generates all interesting patterns of size up to $n$ \cite{pommerening-et-al-ijcai2013} and yields the heuristic \hsys{n}. \sysscp{} exhaustively generates interesting patterns and keeps only those that can increase the estimate of the saturated cost partitioning heuristic \cite{seipp-ijcai2019}; we denote the resulting heuristic \hsysscp{}.

The fifth baseline takes the patterns produced by \hevolved{} and, in each pattern, replaces every non-goal atom with a uniformly random non-goal fluent atom of the task; goal atoms and pattern sizes are preserved. This isolates the contribution of the evolutionary search: if the atom selection in our patterns were no better than random, the resulting heuristic \hrand{} would perform comparably to \hevolved{}.

\subsection{Evolutionary Process}

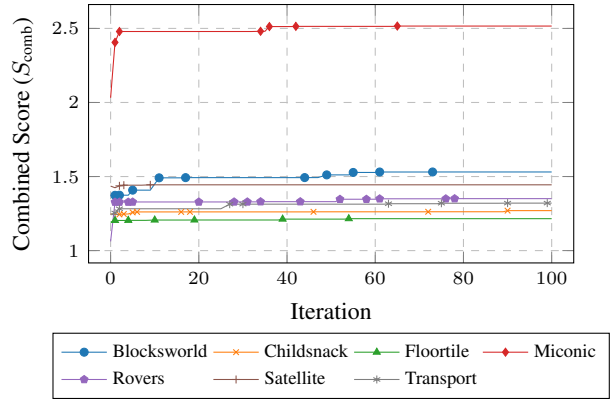
\begin{figure}[tb]
    \definecolor{tab10blue}{HTML}{1f77b4}
\definecolor{tab10orange}{HTML}{ff7f0e}
\definecolor{tab10green}{HTML}{2ca02c}
\definecolor{tab10red}{HTML}{d62728}
\definecolor{tab10purple}{HTML}{9467bd}
\definecolor{tab10brown}{HTML}{8c564b}
\definecolor{tab10pink}{HTML}{e377c2}
\definecolor{tab10gray}{HTML}{7f7f7f}
\definecolor{tab10olive}{HTML}{bcbd22}
\definecolor{tab10cyan}{HTML}{17becf}
\begin{tikzpicture}
    \begin{axis}[
            width=8cm, height=5cm,
            xlabel={\footnotesize Iteration},
            ylabel={\footnotesize Combined Score ($S_{\text{comb}}$)},
            xmin=-5, xmax=105,
            xtick={0,20,40,60,80,100},
            tick label style={font=\scriptsize},
            grid=both,
            grid style=dashed,
            legend style={at={(0.5,-0.27)}, anchor=north, legend columns=4, font=\scriptsize, legend cell align=left},
        ]
        \addplot[color=tab10blue, mark=none, forget plot, restrict expr to domain={\thisrow{domain_idx}}{0:0}]
        table[x=iteration, y=score, col sep=comma] {data/v13_evolution_plot.csv};
        \addplot[color=tab10blue, only marks, mark=*, mark size=1.5pt, forget plot, restrict expr to domain={\thisrow{domain_idx}}{0:0}]
        table[x=iteration, y=mark_score, col sep=comma] {data/v13_evolution_plot.csv};
        \addlegendimage{color=tab10blue,  mark=*, mark size=1.5pt}
        \addlegendentry{Blocksworld}
        \addplot[color=tab10orange, mark=none, forget plot, restrict expr to domain={\thisrow{domain_idx}}{1:1}]
        table[x=iteration, y=score, col sep=comma] {data/v13_evolution_plot.csv};
        \addplot[color=tab10orange, only marks, mark=x, mark size=1.5pt, forget plot, restrict expr to domain={\thisrow{domain_idx}}{1:1}]
        table[x=iteration, y=mark_score, col sep=comma] {data/v13_evolution_plot.csv};
        \addlegendimage{color=tab10orange,  mark=x, mark size=1.5pt}
        \addlegendentry{Childsnack}
        \addplot[color=tab10green, mark=none, forget plot, restrict expr to domain={\thisrow{domain_idx}}{2:2}]
        table[x=iteration, y=score, col sep=comma] {data/v13_evolution_plot.csv};
        \addplot[color=tab10green, only marks, mark=triangle*, mark size=1.5pt, forget plot, restrict expr to domain={\thisrow{domain_idx}}{2:2}]
        table[x=iteration, y=mark_score, col sep=comma] {data/v13_evolution_plot.csv};
        \addlegendimage{color=tab10green,  mark=triangle*, mark size=1.5pt}
        \addlegendentry{Floortile}
        \addplot[color=tab10red, mark=none, forget plot, restrict expr to domain={\thisrow{domain_idx}}{3:3}]
        table[x=iteration, y=score, col sep=comma] {data/v13_evolution_plot.csv};
        \addplot[color=tab10red, only marks, mark=diamond*, mark size=1.5pt, forget plot, restrict expr to domain={\thisrow{domain_idx}}{3:3}]
        table[x=iteration, y=mark_score, col sep=comma] {data/v13_evolution_plot.csv};
        \addlegendimage{color=tab10red,  mark=diamond*, mark size=1.5pt}
        \addlegendentry{Miconic}
        \addplot[color=tab10purple, mark=none, forget plot, restrict expr to domain={\thisrow{domain_idx}}{4:4}]
        table[x=iteration, y=score, col sep=comma] {data/v13_evolution_plot.csv};
        \addplot[color=tab10purple, only marks, mark=pentagon*, mark size=1.5pt, forget plot, restrict expr to domain={\thisrow{domain_idx}}{4:4}]
        table[x=iteration, y=mark_score, col sep=comma] {data/v13_evolution_plot.csv};
        \addlegendimage{color=tab10purple,  mark=pentagon*, mark size=1.5pt}
        \addlegendentry{Rovers}
        \addplot[color=tab10brown, mark=none, forget plot, restrict expr to domain={\thisrow{domain_idx}}{5:5}]
        table[x=iteration, y=score, col sep=comma] {data/v13_evolution_plot.csv};
        \addplot[color=tab10brown, only marks, mark=+, mark size=1.5pt, forget plot, restrict expr to domain={\thisrow{domain_idx}}{5:5}]
        table[x=iteration, y=mark_score, col sep=comma] {data/v13_evolution_plot.csv};
        \addlegendimage{color=tab10brown,  mark=+, mark size=1.5pt}
        \addlegendentry{Satellite}
        \addplot[color=tab10gray, mark=none, forget plot, restrict expr to domain={\thisrow{domain_idx}}{7:7}]
        table[x=iteration, y=score, col sep=comma] {data/v13_evolution_plot.csv};
        \addplot[color=tab10gray, only marks, mark=asterisk, mark size=1.5pt, forget plot, restrict expr to domain={\thisrow{domain_idx}}{7:7}]
        table[x=iteration, y=mark_score, col sep=comma] {data/v13_evolution_plot.csv};
        \addlegendimage{color=tab10gray,  mark=asterisk, mark size=1.5pt}
        \addlegendentry{Transport}
    \end{axis}
\end{tikzpicture}
    \caption{Combined score per iteration for all domains. Markers indicate iterations where a new
        best score was achieved. Most improvement happens in the first five iterations, but
        refinement continues throughout the 100 iterations in most domains. For each domain, the
        point at iteration 0 shows the score of the patterns generated by \sys{1}.}
    \label{plot:score_over_time}
\end{figure}

Figure~\ref{plot:score_over_time} shows how the combined score of the best program evolves over 100 iterations for each domain.
In all domains, the bulk of the improvement occurs within the first five iterations, as the LLM quickly converges from the initial singleton baseline to a domain-tailored strategy.
Evolution, however, keeps refining the program beyond this initial phase.
Satellite stands out: its best generator appears already at iteration 9, suggesting that effective pattern structure is easy to discover here.
For every other domain, the best generator appears at or after iteration 50, with the latest breakthroughs in Childsnack and Transport at iterations 90 and 99, respectively.
These domains require sustained exploration to refine their pattern structure.

Overall, the absolute score improvements range from $+0.009$ in Satellite to $+0.484$ in Miconic, confirming that evolution contributes meaningfully in most domains, even when diminishing returns set in early for simpler ones.

\subsection{Results}
\subsubsection{Coverage}
Table~\ref{tab:results} shows per-domain coverage. \hevolved{} matches or exceeds the best systematic baseline in five of the seven domains: it is uniquely best on Childsnack and Transport, and tied with the strongest baseline on Blocksworld, Rovers and Satellite. The two exceptions are Floortile and Miconic, where systematic enumeration of small patterns dominates and our approach fails to discover comparably effective larger patterns. Pairwise per-domain, \hevolved{} ties or beats \hsysthree{} on six of seven domains (two wins, four ties, one loss) and \hsysscp{} on five (three wins, two ties, two losses); the losses concentrate in Floortile and Miconic.

Aggregating across domains, \hevolved{} solves 61 of 210 tasks against 75 for \hsysthree{} and 74 for \hsysscp{}, but the entire gap stems from Miconic, where \hevolved{} solves only 4 tasks while the two strongest baselines solve 20 and 19. Excluding Miconic, the totals reverse to 57 tasks for \hevolved{} against 55 for both \hsysthree{} and \hsysscp{}.

The random ablation \hrand{} solves 41 tasks: it ties \hevolved{} only on Floortile and Rovers, loses on every other domain, and never beats it. This confirms that the evolutionary process identifies meaningful structural patterns rather than merely selecting atoms at random.

\begin{table}[tb]
  \centering
  \resizebox{\columnwidth}{!}{%
    \begin{tabular}{@{}l r r r r r r@{}}
      \toprule
                                 & \multicolumn{4}{c}{Baselines}                  & \multicolumn{2}{c}{Ours}                                                                                                                                                                                                          \\
      \cmidrule(lr){2-5} \cmidrule(l){6-7}
      Domain                     & \hsysone & \hsystwo & \hsysthree & \hsysscp & \hrand & \hevolved \\
      \midrule
      Blocksworld \numtasks{30}  & 12                                             & \textbf{16}                                    & \textbf{16}                                    & \textbf{16}                                       & 12                                             & \textbf{16}                     \\
      Childsnack \numtasks{30}   & 4                                              & 4                                              & 4                                              & 4                                                 & 4                                              & \textbf{5}                      \\
      Floortile \numtasks{30}    & 4                                              & 5                                              & 5                                              & \textbf{7}                                        & 5                                              & 5                               \\
      Miconic \numtasks{30}      & 3                                              & 6                                              & \textbf{20}                                    & 19                                                & 3                                              & 4                               \\
      Rovers \numtasks{30}       & \textbf{2}                                     & \textbf{2}                                     & \textbf{2}                                     & \textbf{2}                                        & \textbf{2}                                     & \textbf{2}                      \\
      Satellite \numtasks{30}    & 5                                              & 15                                             & \textbf{16}                                    & 15                                                & 5                                              & \textbf{16}                     \\
      Transport \numtasks{30}    & 10                                             & 10                                             & 12                                             & 11                                                & 10                                             & \textbf{13}                     \\
      \midrule
      Sum \numtasks{210} & 40                                            & 58                                             & \textbf{75}                                    & 74                                                & 41                                             & 61                              \\
      \bottomrule
    \end{tabular}%
  }
  \caption{Per-domain and overall coverage scores for each heuristic with the
    the best scores highlighted in bold.
    }
  \label{tab:results}
\end{table}

\subsubsection{Expansions Before the Last $f$-Layer}
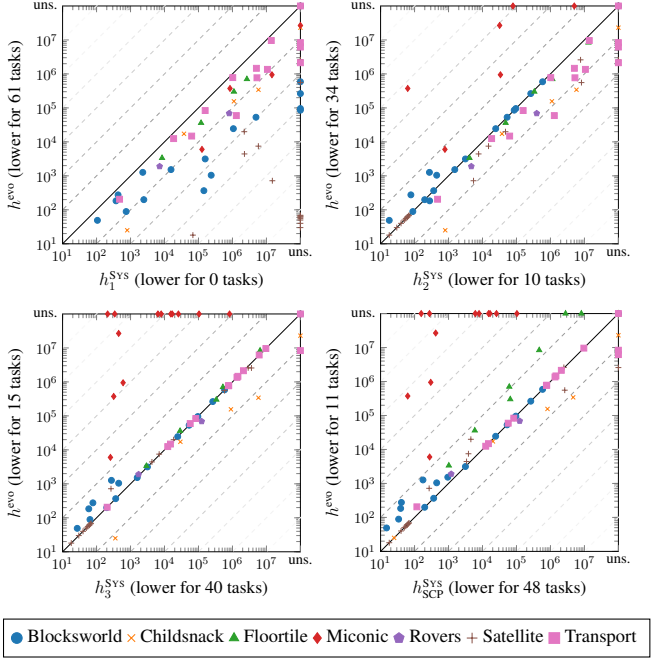
\begin{figure}[h]
    \centering
    \makeatletter
\@ifundefined{scatterlegenddefined}{%
    \gdef\scatterlegenddefined{1}%
    \begin{tikzpicture}
        \begin{axis}[
                hide axis,
                xmin=1, xmax=2, ymin=1, ymax=2,
                clip=true,
                legend to name=scatterlegend,
                legend cell align=left,
                legend columns=7,
                legend style={font=\scriptsize},
            ]
            \addplot+[only marks] coordinates {(0,0)}; \addlegendentry{Blocksworld}
            \addplot+[only marks] coordinates {(0,0)}; \addlegendentry{Childsnack}
            \addplot+[only marks] coordinates {(0,0)}; \addlegendentry{Floortile}
            \addplot+[only marks] coordinates {(0,0)}; \addlegendentry{Miconic}
            \addplot+[only marks] coordinates {(0,0)}; \addlegendentry{Rovers}
            \addplot+[only marks] coordinates {(0,0)}; \addlegendentry{Satellite}
            \addplot+[only marks] coordinates {(0,0)}; \addlegendentry{Transport}
        \end{axis}
    \end{tikzpicture}%
}{}
\makeatother%
    \resizebox{\columnwidth}{!}{%
        \begin{tikzpicture}
      \begin{axis}[
                  xmin=10,
                  xmax=100000000,
                  xtick={1,10,100,1000,10000,100000,1000000,10000000},
                  extra x ticks={0.25,100000000},
                  extra x tick labels={0,uns.},
                  xmode=log,
                  ymin=10,
                  ymax=100000000,
                  ytick={1,10,100,1000,10000,100000,1000000,10000000},
                  extra y ticks={0.25,100000000},
                  extra y tick labels={0,uns.},
                  ymode=log,
                  axis equal image,
                  clip mode=global,
                  xlabel={\hsysone{} (lower for 0 tasks)},
                  ylabel={\hevolved{} (lower for 61 tasks)},
                  label style={font=\Large}]
            \addplot[color=black!3, dashed, forget plot] coordinates {(0.1, 100000) (1000, 1000000000)};
            \addplot[color=black!5, dashed, forget plot] coordinates {(0.1, 10000) (10000, 1000000000)};
            \addplot[color=black!10, dashed, forget plot] coordinates {(0.1, 1000) (100000, 1000000000)};
            \addplot[color=black!20, dashed, forget plot] coordinates {(0.1, 100) (1000000, 1000000000)};
            \addplot[color=black!30, dashed, forget plot] coordinates {(0.1, 10) (10000000, 1000000000)};
            \addplot[color=black!40, dashed, forget plot] coordinates {(0.1, 1) (100000000, 1000000000)};
            \addplot[color=black!40, dashed, forget plot] coordinates {(1, 0.1) (1000000000, 100000000)};
            \addplot[color=black!30, dashed, forget plot] coordinates {(10, 0.1) (1000000000, 10000000)};
            \addplot[color=black!20, dashed, forget plot] coordinates {(100, 0.1) (1000000000, 1000000)};
            \addplot[color=black!10, dashed, forget plot] coordinates {(1000, 0.1) (1000000000, 100000)};
            \addplot[color=black!5, dashed, forget plot] coordinates {(10000, 0.1) (1000000000, 10000)};
            \addplot[color=black!3, dashed, forget plot] coordinates {(100000, 0.1) (1000000000, 1000)};
            \addplot+[only marks, restrict expr to domain={\thisrow{cat}}{0:0}] table[col sep=comma, x=x, y=y]{data/combined-2026-05-05-scatter-sys1-vs-evolved-patterns-expansions_until_last_jump.csv};
            \addplot+[only marks, restrict expr to domain={\thisrow{cat}}{1:1}] table[col sep=comma, x=x, y=y]{data/combined-2026-05-05-scatter-sys1-vs-evolved-patterns-expansions_until_last_jump.csv};
            \addplot+[only marks, restrict expr to domain={\thisrow{cat}}{2:2}] table[col sep=comma, x=x, y=y]{data/combined-2026-05-05-scatter-sys1-vs-evolved-patterns-expansions_until_last_jump.csv};
            \addplot+[only marks, restrict expr to domain={\thisrow{cat}}{3:3}] table[col sep=comma, x=x, y=y]{data/combined-2026-05-05-scatter-sys1-vs-evolved-patterns-expansions_until_last_jump.csv};
            \addplot+[only marks, restrict expr to domain={\thisrow{cat}}{4:4}] table[col sep=comma, x=x, y=y]{data/combined-2026-05-05-scatter-sys1-vs-evolved-patterns-expansions_until_last_jump.csv};
            \addplot+[only marks, restrict expr to domain={\thisrow{cat}}{5:5}] table[col sep=comma, x=x, y=y]{data/combined-2026-05-05-scatter-sys1-vs-evolved-patterns-expansions_until_last_jump.csv};
            \addplot+[only marks, restrict expr to domain={\thisrow{cat}}{7:7}] table[col sep=comma, x=x, y=y]{data/combined-2026-05-05-scatter-sys1-vs-evolved-patterns-expansions_until_last_jump.csv};
            \draw[color=black] (axis cs:1e-70,1e-70) -- (axis cs:1e70,1e70);
      \end{axis}
\end{tikzpicture}%
        \begin{tikzpicture}
      \begin{axis}[
                  xmin=10,
                  xmax=100000000,
                  xtick={1,10,100,1000,10000,100000,1000000,10000000},
                  extra x ticks={0.25,100000000},
                  extra x tick labels={0,uns.},
                  xmode=log,
                  ymin=10,
                  ymax=100000000,
                  ytick={1,10,100,1000,10000,100000,1000000,10000000},
                  extra y ticks={0.25,100000000},
                  extra y tick labels={0,uns.},
                  ymode=log,
                  axis equal image,
                  clip mode=global,
                  xlabel={\hsystwo{} (lower for 10 tasks)},
                  ylabel={\hevolved{} (lower for 34 tasks)},
                  label style={font=\Large}]
            \addplot[color=black!3, dashed, forget plot] coordinates {(0.1, 100000) (1000, 1000000000)};
            \addplot[color=black!5, dashed, forget plot] coordinates {(0.1, 10000) (10000, 1000000000)};
            \addplot[color=black!10, dashed, forget plot] coordinates {(0.1, 1000) (100000, 1000000000)};
            \addplot[color=black!20, dashed, forget plot] coordinates {(0.1, 100) (1000000, 1000000000)};
            \addplot[color=black!30, dashed, forget plot] coordinates {(0.1, 10) (10000000, 1000000000)};
            \addplot[color=black!40, dashed, forget plot] coordinates {(0.1, 1) (100000000, 1000000000)};
            \addplot[color=black!40, dashed, forget plot] coordinates {(1, 0.1) (1000000000, 100000000)};
            \addplot[color=black!30, dashed, forget plot] coordinates {(10, 0.1) (1000000000, 10000000)};
            \addplot[color=black!20, dashed, forget plot] coordinates {(100, 0.1) (1000000000, 1000000)};
            \addplot[color=black!10, dashed, forget plot] coordinates {(1000, 0.1) (1000000000, 100000)};
            \addplot[color=black!5, dashed, forget plot] coordinates {(10000, 0.1) (1000000000, 10000)};
            \addplot[color=black!3, dashed, forget plot] coordinates {(100000, 0.1) (1000000000, 1000)};
            \addplot+[only marks, restrict expr to domain={\thisrow{cat}}{0:0}] table[col sep=comma, x=x, y=y]{data/combined-2026-05-05-scatter-sys2-vs-evolved-patterns-expansions_until_last_jump.csv};
            \addplot+[only marks, restrict expr to domain={\thisrow{cat}}{1:1}] table[col sep=comma, x=x, y=y]{data/combined-2026-05-05-scatter-sys2-vs-evolved-patterns-expansions_until_last_jump.csv};
            \addplot+[only marks, restrict expr to domain={\thisrow{cat}}{2:2}] table[col sep=comma, x=x, y=y]{data/combined-2026-05-05-scatter-sys2-vs-evolved-patterns-expansions_until_last_jump.csv};
            \addplot+[only marks, restrict expr to domain={\thisrow{cat}}{3:3}] table[col sep=comma, x=x, y=y]{data/combined-2026-05-05-scatter-sys2-vs-evolved-patterns-expansions_until_last_jump.csv};
            \addplot+[only marks, restrict expr to domain={\thisrow{cat}}{4:4}] table[col sep=comma, x=x, y=y]{data/combined-2026-05-05-scatter-sys2-vs-evolved-patterns-expansions_until_last_jump.csv};
            \addplot+[only marks, restrict expr to domain={\thisrow{cat}}{5:5}] table[col sep=comma, x=x, y=y]{data/combined-2026-05-05-scatter-sys2-vs-evolved-patterns-expansions_until_last_jump.csv};
            \addplot+[only marks, restrict expr to domain={\thisrow{cat}}{7:7}] table[col sep=comma, x=x, y=y]{data/combined-2026-05-05-scatter-sys2-vs-evolved-patterns-expansions_until_last_jump.csv};
            \draw[color=black] (axis cs:1e-70,1e-70) -- (axis cs:1e70,1e70);
      \end{axis}
\end{tikzpicture}%
    }\\[0.3em]
    \resizebox{\columnwidth}{!}{%
        \begin{tikzpicture}
      \begin{axis}[
                  xmin=10,
                  xmax=100000000,
                  xtick={1,10,100,1000,10000,100000,1000000,10000000},
                  extra x ticks={0.25,100000000},
                  extra x tick labels={0,uns.},
                  xmode=log,
                  ymin=10,
                  ymax=100000000,
                  ytick={1,10,100,1000,10000,100000,1000000,10000000},
                  extra y ticks={0.25,100000000},
                  extra y tick labels={0,uns.},
                  ymode=log,
                  axis equal image,
                  clip mode=global,
                  xlabel={\hsysthree{} (lower for 40 tasks)},
                  ylabel={\hevolved{} (lower for 15 tasks)},
                  label style={font=\Large}]
            \addplot[color=black!3, dashed, forget plot] coordinates {(0.1, 100000) (1000, 1000000000)};
            \addplot[color=black!5, dashed, forget plot] coordinates {(0.1, 10000) (10000, 1000000000)};
            \addplot[color=black!10, dashed, forget plot] coordinates {(0.1, 1000) (100000, 1000000000)};
            \addplot[color=black!20, dashed, forget plot] coordinates {(0.1, 100) (1000000, 1000000000)};
            \addplot[color=black!30, dashed, forget plot] coordinates {(0.1, 10) (10000000, 1000000000)};
            \addplot[color=black!40, dashed, forget plot] coordinates {(0.1, 1) (100000000, 1000000000)};
            \addplot[color=black!40, dashed, forget plot] coordinates {(1, 0.1) (1000000000, 100000000)};
            \addplot[color=black!30, dashed, forget plot] coordinates {(10, 0.1) (1000000000, 10000000)};
            \addplot[color=black!20, dashed, forget plot] coordinates {(100, 0.1) (1000000000, 1000000)};
            \addplot[color=black!10, dashed, forget plot] coordinates {(1000, 0.1) (1000000000, 100000)};
            \addplot[color=black!5, dashed, forget plot] coordinates {(10000, 0.1) (1000000000, 10000)};
            \addplot[color=black!3, dashed, forget plot] coordinates {(100000, 0.1) (1000000000, 1000)};
            \addplot+[only marks, restrict expr to domain={\thisrow{cat}}{0:0}] table[col sep=comma, x=x, y=y]{data/combined-2026-05-05-scatter-sys3-vs-evolved-patterns-expansions_until_last_jump.csv};
            \addplot+[only marks, restrict expr to domain={\thisrow{cat}}{1:1}] table[col sep=comma, x=x, y=y]{data/combined-2026-05-05-scatter-sys3-vs-evolved-patterns-expansions_until_last_jump.csv};
            \addplot+[only marks, restrict expr to domain={\thisrow{cat}}{2:2}] table[col sep=comma, x=x, y=y]{data/combined-2026-05-05-scatter-sys3-vs-evolved-patterns-expansions_until_last_jump.csv};
            \addplot+[only marks, restrict expr to domain={\thisrow{cat}}{3:3}] table[col sep=comma, x=x, y=y]{data/combined-2026-05-05-scatter-sys3-vs-evolved-patterns-expansions_until_last_jump.csv};
            \addplot+[only marks, restrict expr to domain={\thisrow{cat}}{4:4}] table[col sep=comma, x=x, y=y]{data/combined-2026-05-05-scatter-sys3-vs-evolved-patterns-expansions_until_last_jump.csv};
            \addplot+[only marks, restrict expr to domain={\thisrow{cat}}{5:5}] table[col sep=comma, x=x, y=y]{data/combined-2026-05-05-scatter-sys3-vs-evolved-patterns-expansions_until_last_jump.csv};
            \addplot+[only marks, restrict expr to domain={\thisrow{cat}}{7:7}] table[col sep=comma, x=x, y=y]{data/combined-2026-05-05-scatter-sys3-vs-evolved-patterns-expansions_until_last_jump.csv};
            \draw[color=black] (axis cs:1e-70,1e-70) -- (axis cs:1e70,1e70);
      \end{axis}
\end{tikzpicture}%
        \begin{tikzpicture}
      \begin{axis}[
                  xmin=10,
                  xmax=100000000,
                  xtick={1,10,100,1000,10000,100000,1000000,10000000},
                  extra x ticks={0.25,100000000},
                  extra x tick labels={0,uns.},
                  xmode=log,
                  ymin=10,
                  ymax=100000000,
                  ytick={1,10,100,1000,10000,100000,1000000,10000000},
                  extra y ticks={0.25,100000000},
                  extra y tick labels={0,uns.},
                  ymode=log,
                  axis equal image,
                  clip mode=global,
                  xlabel={\hsysscp{} (lower for 48 tasks)},
                  ylabel={\hevolved{} (lower for 11 tasks)},
                  label style={font=\Large}]
            \addplot[color=black!3, dashed, forget plot] coordinates {(0.1, 100000) (1000, 1000000000)};
            \addplot[color=black!5, dashed, forget plot] coordinates {(0.1, 10000) (10000, 1000000000)};
            \addplot[color=black!10, dashed, forget plot] coordinates {(0.1, 1000) (100000, 1000000000)};
            \addplot[color=black!20, dashed, forget plot] coordinates {(0.1, 100) (1000000, 1000000000)};
            \addplot[color=black!30, dashed, forget plot] coordinates {(0.1, 10) (10000000, 1000000000)};
            \addplot[color=black!40, dashed, forget plot] coordinates {(0.1, 1) (100000000, 1000000000)};
            \addplot[color=black!40, dashed, forget plot] coordinates {(1, 0.1) (1000000000, 100000000)};
            \addplot[color=black!30, dashed, forget plot] coordinates {(10, 0.1) (1000000000, 10000000)};
            \addplot[color=black!20, dashed, forget plot] coordinates {(100, 0.1) (1000000000, 1000000)};
            \addplot[color=black!10, dashed, forget plot] coordinates {(1000, 0.1) (1000000000, 100000)};
            \addplot[color=black!5, dashed, forget plot] coordinates {(10000, 0.1) (1000000000, 10000)};
            \addplot[color=black!3, dashed, forget plot] coordinates {(100000, 0.1) (1000000000, 1000)};
            \addplot+[only marks, restrict expr to domain={\thisrow{cat}}{0:0}] table[col sep=comma, x=x, y=y]{data/combined-2026-05-05-scatter-sys-scp-vs-evolved-patterns-expansions_until_last_jump.csv};
            \addplot+[only marks, restrict expr to domain={\thisrow{cat}}{1:1}] table[col sep=comma, x=x, y=y]{data/combined-2026-05-05-scatter-sys-scp-vs-evolved-patterns-expansions_until_last_jump.csv};
            \addplot+[only marks, restrict expr to domain={\thisrow{cat}}{2:2}] table[col sep=comma, x=x, y=y]{data/combined-2026-05-05-scatter-sys-scp-vs-evolved-patterns-expansions_until_last_jump.csv};
            \addplot+[only marks, restrict expr to domain={\thisrow{cat}}{3:3}] table[col sep=comma, x=x, y=y]{data/combined-2026-05-05-scatter-sys-scp-vs-evolved-patterns-expansions_until_last_jump.csv};
            \addplot+[only marks, restrict expr to domain={\thisrow{cat}}{4:4}] table[col sep=comma, x=x, y=y]{data/combined-2026-05-05-scatter-sys-scp-vs-evolved-patterns-expansions_until_last_jump.csv};
            \addplot+[only marks, restrict expr to domain={\thisrow{cat}}{5:5}] table[col sep=comma, x=x, y=y]{data/combined-2026-05-05-scatter-sys-scp-vs-evolved-patterns-expansions_until_last_jump.csv};
            \addplot+[only marks, restrict expr to domain={\thisrow{cat}}{7:7}] table[col sep=comma, x=x, y=y]{data/combined-2026-05-05-scatter-sys-scp-vs-evolved-patterns-expansions_until_last_jump.csv};
            \draw[color=black] (axis cs:1e-70,1e-70) -- (axis cs:1e70,1e70);
      \end{axis}
\end{tikzpicture}%
    }%
    \par\vspace{0.5em}%
    {\centering\pgfplotslegendfromname{scatterlegend}\par}%
    \caption{Expansions until last $f$-layer: \hevolved{} vs.\ each baseline. Each
        point is one task; points below the diagonal indicate tasks where
        \hevolved{} requires fewer expansions.}
    \label{plot:expansions}
\end{figure}

Figure~\ref{plot:expansions} shows guidance quality, measured in expansions before the last $f$-layer.
\hevolved{} dominates \hsysone{} on all tasks solved by both heuristics, and needs fewer expansions than \hsystwo{} on 34 tasks out of 44.
The balance shifts in favor of the systematic baselines at larger pattern sizes: against \hsysthree{} and \hsysscp{}, \hevolved{} performs fewer expansions on only 15 tasks out of 55 and 11 tasks out of 59, respectively.
This reveals the core trade-off of our synthesized pattern generators: exhaustive systematic enumeration at size 3 yields stronger per-state guidance than our domain-specific collections, but our generators compensate through faster evaluation.

\subsubsection{Total Search Time}
\begin{figure}[h]
    \centering
    \makeatletter
\@ifundefined{scatterlegenddefined}{%
    \gdef\scatterlegenddefined{1}%
    \begin{tikzpicture}
        \begin{axis}[
                hide axis,
                xmin=1, xmax=2, ymin=1, ymax=2,
                clip=true,
                legend to name=scatterlegend,
                legend cell align=left,
                legend columns=7,
                legend style={font=\scriptsize},
            ]
            \addplot+[only marks] coordinates {(0,0)}; \addlegendentry{Blocksworld}
            \addplot+[only marks] coordinates {(0,0)}; \addlegendentry{Childsnack}
            \addplot+[only marks] coordinates {(0,0)}; \addlegendentry{Floortile}
            \addplot+[only marks] coordinates {(0,0)}; \addlegendentry{Miconic}
            \addplot+[only marks] coordinates {(0,0)}; \addlegendentry{Rovers}
            \addplot+[only marks] coordinates {(0,0)}; \addlegendentry{Satellite}
            \addplot+[only marks] coordinates {(0,0)}; \addlegendentry{Transport}
        \end{axis}
    \end{tikzpicture}%
}{}
\makeatother%
    \resizebox{\columnwidth}{!}{%
        \begin{tikzpicture}
      \begin{axis}[
                  xmin=0.1,
                  xmax=1000,
                  xtick={1,10,100},
                  extra x ticks={0.25,1000},
                  extra x tick labels={0,uns.},
                  xmode=log,
                  ymin=0.1,
                  ymax=1000,
                  ytick={1,10,100},
                  extra y ticks={0.25,1000},
                  extra y tick labels={0,uns.},
                  ymode=log,
                  axis equal image,
                  xlabel={\hsysone{} (lower for 6 tasks)},
                  ylabel={\hevolved{} (lower for 52 tasks)},
                  label style={font=\Large}]
            \addplot[color=black!40, dashed, forget plot] coordinates {(0.1, 1.0) (100.0, 1000)};
            \addplot[color=black!40, dashed, forget plot] coordinates {(1.0, 0.1) (1000, 100.0)};
            \addplot[color=black!30, dashed, forget plot] coordinates {(0.1, 10.0) (10.0, 1000)};
            \addplot[color=black!30, dashed, forget plot] coordinates {(10.0, 0.1) (1000, 10.0)};
            \addplot[color=black!20, dashed, forget plot] coordinates {(0.1, 100.0) (1.0, 1000)};
            \addplot[color=black!20, dashed, forget plot] coordinates {(100.0, 0.1) (1000, 1.0)};
            \addplot+[only marks, restrict expr to domain={\thisrow{cat}}{0:0}] table[col sep=comma, x=x, y=y]{data/combined-2026-05-05-scatter-sys1-vs-evolved-patterns-total_time.csv};
            \addplot+[only marks, restrict expr to domain={\thisrow{cat}}{1:1}] table[col sep=comma, x=x, y=y]{data/combined-2026-05-05-scatter-sys1-vs-evolved-patterns-total_time.csv};
            \addplot+[only marks, restrict expr to domain={\thisrow{cat}}{2:2}] table[col sep=comma, x=x, y=y]{data/combined-2026-05-05-scatter-sys1-vs-evolved-patterns-total_time.csv};
            \addplot+[only marks, restrict expr to domain={\thisrow{cat}}{3:3}] table[col sep=comma, x=x, y=y]{data/combined-2026-05-05-scatter-sys1-vs-evolved-patterns-total_time.csv};
            \addplot+[only marks, restrict expr to domain={\thisrow{cat}}{4:4}] table[col sep=comma, x=x, y=y]{data/combined-2026-05-05-scatter-sys1-vs-evolved-patterns-total_time.csv};
            \addplot+[only marks, restrict expr to domain={\thisrow{cat}}{5:5}] table[col sep=comma, x=x, y=y]{data/combined-2026-05-05-scatter-sys1-vs-evolved-patterns-total_time.csv};
            \addplot+[only marks, restrict expr to domain={\thisrow{cat}}{7:7}] table[col sep=comma, x=x, y=y]{data/combined-2026-05-05-scatter-sys1-vs-evolved-patterns-total_time.csv};
            \draw[color=black] (axis cs:1e-70,1e-70) -- (axis cs:1e70,1e70);
      \end{axis}
\end{tikzpicture}%
        \begin{tikzpicture}
      \begin{axis}[
                  xmin=0.1,
                  xmax=1000,
                  xtick={1,10,100},
                  extra x ticks={0.25,1000},
                  extra x tick labels={0,uns.},
                  xmode=log,
                  ymin=0.1,
                  ymax=1000,
                  ytick={1,10,100},
                  extra y ticks={0.25,1000},
                  extra y tick labels={0,uns.},
                  ymode=log,
                  axis equal image,
                  xlabel={\hsystwo{} (lower for 31 tasks)},
                  ylabel={\hevolved{} (lower for 27 tasks)},
                  label style={font=\Large}]
            \addplot[color=black!40, dashed, forget plot] coordinates {(0.1, 1.0) (100.0, 1000)};
            \addplot[color=black!40, dashed, forget plot] coordinates {(1.0, 0.1) (1000, 100.0)};
            \addplot[color=black!30, dashed, forget plot] coordinates {(0.1, 10.0) (10.0, 1000)};
            \addplot[color=black!30, dashed, forget plot] coordinates {(10.0, 0.1) (1000, 10.0)};
            \addplot[color=black!20, dashed, forget plot] coordinates {(0.1, 100.0) (1.0, 1000)};
            \addplot[color=black!20, dashed, forget plot] coordinates {(100.0, 0.1) (1000, 1.0)};
            \addplot+[only marks, restrict expr to domain={\thisrow{cat}}{0:0}] table[col sep=comma, x=x, y=y]{data/combined-2026-05-05-scatter-sys2-vs-evolved-patterns-total_time.csv};
            \addplot+[only marks, restrict expr to domain={\thisrow{cat}}{1:1}] table[col sep=comma, x=x, y=y]{data/combined-2026-05-05-scatter-sys2-vs-evolved-patterns-total_time.csv};
            \addplot+[only marks, restrict expr to domain={\thisrow{cat}}{2:2}] table[col sep=comma, x=x, y=y]{data/combined-2026-05-05-scatter-sys2-vs-evolved-patterns-total_time.csv};
            \addplot+[only marks, restrict expr to domain={\thisrow{cat}}{3:3}] table[col sep=comma, x=x, y=y]{data/combined-2026-05-05-scatter-sys2-vs-evolved-patterns-total_time.csv};
            \addplot+[only marks, restrict expr to domain={\thisrow{cat}}{4:4}] table[col sep=comma, x=x, y=y]{data/combined-2026-05-05-scatter-sys2-vs-evolved-patterns-total_time.csv};
            \addplot+[only marks, restrict expr to domain={\thisrow{cat}}{5:5}] table[col sep=comma, x=x, y=y]{data/combined-2026-05-05-scatter-sys2-vs-evolved-patterns-total_time.csv};
            \addplot+[only marks, restrict expr to domain={\thisrow{cat}}{7:7}] table[col sep=comma, x=x, y=y]{data/combined-2026-05-05-scatter-sys2-vs-evolved-patterns-total_time.csv};
            \draw[color=black] (axis cs:1e-70,1e-70) -- (axis cs:1e70,1e70);
      \end{axis}
\end{tikzpicture}%
    }\\[0.3em]
    \resizebox{\columnwidth}{!}{%
        \begin{tikzpicture}
      \begin{axis}[
                  xmin=0.1,
                  xmax=1000,
                  xtick={1,10,100},
                  extra x ticks={0.25,1000},
                  extra x tick labels={0,uns.},
                  xmode=log,
                  ymin=0.1,
                  ymax=1000,
                  ytick={1,10,100},
                  extra y ticks={0.25,1000},
                  extra y tick labels={0,uns.},
                  ymode=log,
                  axis equal image,
                  xlabel={\hsysthree{} (lower for 22 tasks)},
                  ylabel={\hevolved{} (lower for 54 tasks)},
                  label style={font=\Large}]
            \addplot[color=black!40, dashed, forget plot] coordinates {(0.1, 1.0) (100.0, 1000)};
            \addplot[color=black!40, dashed, forget plot] coordinates {(1.0, 0.1) (1000, 100.0)};
            \addplot[color=black!30, dashed, forget plot] coordinates {(0.1, 10.0) (10.0, 1000)};
            \addplot[color=black!30, dashed, forget plot] coordinates {(10.0, 0.1) (1000, 10.0)};
            \addplot[color=black!20, dashed, forget plot] coordinates {(0.1, 100.0) (1.0, 1000)};
            \addplot[color=black!20, dashed, forget plot] coordinates {(100.0, 0.1) (1000, 1.0)};
            \addplot+[only marks, restrict expr to domain={\thisrow{cat}}{0:0}] table[col sep=comma, x=x, y=y]{data/combined-2026-05-05-scatter-sys3-vs-evolved-patterns-total_time.csv};
            \addplot+[only marks, restrict expr to domain={\thisrow{cat}}{1:1}] table[col sep=comma, x=x, y=y]{data/combined-2026-05-05-scatter-sys3-vs-evolved-patterns-total_time.csv};
            \addplot+[only marks, restrict expr to domain={\thisrow{cat}}{2:2}] table[col sep=comma, x=x, y=y]{data/combined-2026-05-05-scatter-sys3-vs-evolved-patterns-total_time.csv};
            \addplot+[only marks, restrict expr to domain={\thisrow{cat}}{3:3}] table[col sep=comma, x=x, y=y]{data/combined-2026-05-05-scatter-sys3-vs-evolved-patterns-total_time.csv};
            \addplot+[only marks, restrict expr to domain={\thisrow{cat}}{4:4}] table[col sep=comma, x=x, y=y]{data/combined-2026-05-05-scatter-sys3-vs-evolved-patterns-total_time.csv};
            \addplot+[only marks, restrict expr to domain={\thisrow{cat}}{5:5}] table[col sep=comma, x=x, y=y]{data/combined-2026-05-05-scatter-sys3-vs-evolved-patterns-total_time.csv};
            \addplot+[only marks, restrict expr to domain={\thisrow{cat}}{7:7}] table[col sep=comma, x=x, y=y]{data/combined-2026-05-05-scatter-sys3-vs-evolved-patterns-total_time.csv};
            \draw[color=black] (axis cs:1e-70,1e-70) -- (axis cs:1e70,1e70);
      \end{axis}
\end{tikzpicture}%
        \begin{tikzpicture}
      \begin{axis}[
                  xmin=0.1,
                  xmax=1000,
                  xtick={1,10,100},
                  extra x ticks={0.25,1000},
                  extra x tick labels={0,uns.},
                  xmode=log,
                  ymin=0.1,
                  ymax=1000,
                  ytick={1,10,100},
                  extra y ticks={0.25,1000},
                  extra y tick labels={0,uns.},
                  ymode=log,
                  axis equal image,
                  xlabel={\hsysscp{} (lower for 18 tasks)},
                  ylabel={\hevolved{} (lower for 60 tasks)},
                  label style={font=\Large}]
            \addplot[color=black!40, dashed, forget plot] coordinates {(0.1, 1.0) (100.0, 1000)};
            \addplot[color=black!40, dashed, forget plot] coordinates {(1.0, 0.1) (1000, 100.0)};
            \addplot[color=black!30, dashed, forget plot] coordinates {(0.1, 10.0) (10.0, 1000)};
            \addplot[color=black!30, dashed, forget plot] coordinates {(10.0, 0.1) (1000, 10.0)};
            \addplot[color=black!20, dashed, forget plot] coordinates {(0.1, 100.0) (1.0, 1000)};
            \addplot[color=black!20, dashed, forget plot] coordinates {(100.0, 0.1) (1000, 1.0)};
            \addplot+[only marks, restrict expr to domain={\thisrow{cat}}{0:0}] table[col sep=comma, x=x, y=y]{data/combined-2026-05-05-scatter-sys-scp-vs-evolved-patterns-total_time.csv};
            \addplot+[only marks, restrict expr to domain={\thisrow{cat}}{1:1}] table[col sep=comma, x=x, y=y]{data/combined-2026-05-05-scatter-sys-scp-vs-evolved-patterns-total_time.csv};
            \addplot+[only marks, restrict expr to domain={\thisrow{cat}}{2:2}] table[col sep=comma, x=x, y=y]{data/combined-2026-05-05-scatter-sys-scp-vs-evolved-patterns-total_time.csv};
            \addplot+[only marks, restrict expr to domain={\thisrow{cat}}{3:3}] table[col sep=comma, x=x, y=y]{data/combined-2026-05-05-scatter-sys-scp-vs-evolved-patterns-total_time.csv};
            \addplot+[only marks, restrict expr to domain={\thisrow{cat}}{4:4}] table[col sep=comma, x=x, y=y]{data/combined-2026-05-05-scatter-sys-scp-vs-evolved-patterns-total_time.csv};
            \addplot+[only marks, restrict expr to domain={\thisrow{cat}}{5:5}] table[col sep=comma, x=x, y=y]{data/combined-2026-05-05-scatter-sys-scp-vs-evolved-patterns-total_time.csv};
            \addplot+[only marks, restrict expr to domain={\thisrow{cat}}{7:7}] table[col sep=comma, x=x, y=y]{data/combined-2026-05-05-scatter-sys-scp-vs-evolved-patterns-total_time.csv};
            \draw[color=black] (axis cs:1e-70,1e-70) -- (axis cs:1e70,1e70);
      \end{axis}
\end{tikzpicture}%
    }%
    \par\vspace{0.5em}%
    {\centering\pgfplotslegendfromname{scatterlegend}\par}%
    \caption{Total search time: \hevolved{} vs.\ each baseline. Points below the
        diagonal indicate tasks where \hevolved{} is faster.}
    \label{plot:time}
\end{figure}
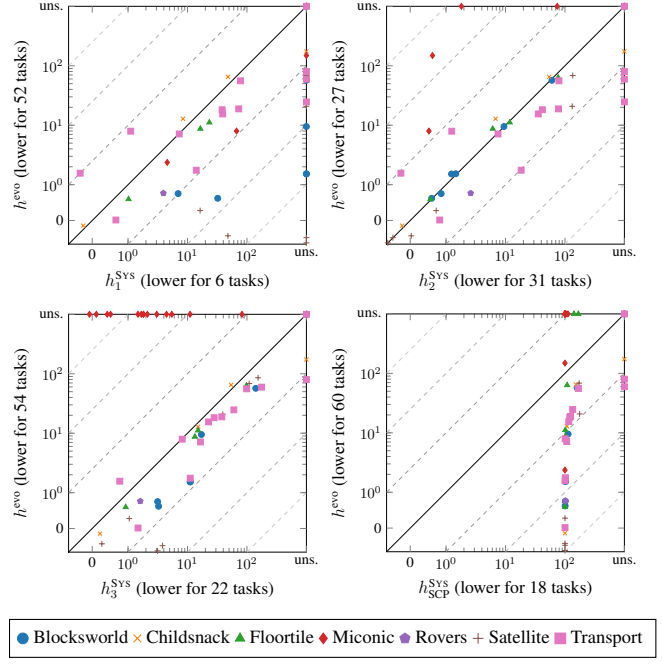

Figure~\ref{plot:time} shows total search time, which includes the computation of all PDBs and the time spent in search.
Although \hevolved{} provides weaker per-state guidance than the strongest baselines, it is faster overall in most comparisons.
It beats \hsysone{} on 52 tasks out of 58, as the singleton patterns of \hsysone{} are efficient to evaluate but provide little guidance.
Against \hsysthree{}, \hevolved{} is faster on 54 tasks out of 76, and against \hsysscp{}, on 60 tasks out of 78.

\hsysscp{} spends up to 100\,s per task on pattern generation, which is its bottleneck.
\hevolved{} shifts that cost to a one-time offline evolution; at test time, the learned generator produces a pattern collection in negligible time.

\subsubsection{Time per Evaluation}
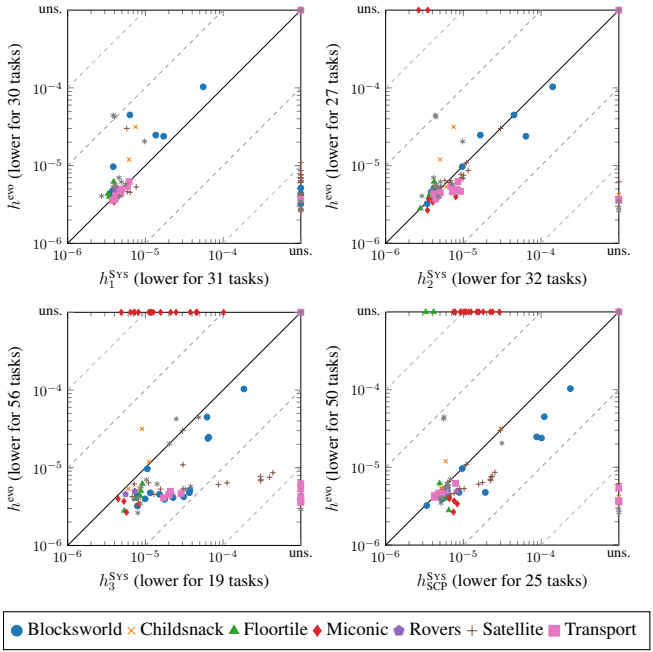
\begin{figure}[h]
    \centering
    \makeatletter
\@ifundefined{scatterlegenddefined}{%
    \gdef\scatterlegenddefined{1}%
    \begin{tikzpicture}
        \begin{axis}[
                hide axis,
                xmin=1, xmax=2, ymin=1, ymax=2,
                clip=true,
                legend to name=scatterlegend,
                legend cell align=left,
                legend columns=7,
                legend style={font=\scriptsize},
            ]
            \addplot+[only marks] coordinates {(0,0)}; \addlegendentry{Blocksworld}
            \addplot+[only marks] coordinates {(0,0)}; \addlegendentry{Childsnack}
            \addplot+[only marks] coordinates {(0,0)}; \addlegendentry{Floortile}
            \addplot+[only marks] coordinates {(0,0)}; \addlegendentry{Miconic}
            \addplot+[only marks] coordinates {(0,0)}; \addlegendentry{Rovers}
            \addplot+[only marks] coordinates {(0,0)}; \addlegendentry{Satellite}
            \addplot+[only marks] coordinates {(0,0)}; \addlegendentry{Transport}
        \end{axis}
    \end{tikzpicture}%
}{}
\makeatother%
    \resizebox{\columnwidth}{!}{%
        \begin{tikzpicture}
      \begin{axis}[
                  xmin=0.000001,
                  xmax=0.001,
                  xtick={0.000001,0.00001,0.0001},
                  extra x ticks={0.001},
                  extra x tick labels={uns.},
                  xmode=log,
                  ymin=0.000001,
                  ymax=0.001,
                  ytick={0.000001,0.00001,0.0001},
                  extra y ticks={0.001},
                  extra y tick labels={uns.},
                  ymode=log,
                  axis equal image,
                  xlabel={\hsysone{} (lower for 31 tasks)},
                  ylabel={\hevolved{} (lower for 30 tasks)},
                  label style={font=\Large}]
            \addplot[color=black!3, dashed, forget plot] coordinates {(0.000001, 1) (100, 100000000)};
            \addplot[color=black!5, dashed, forget plot] coordinates {(0.000001, 0.1) (1000, 100000000)};
            \addplot[color=black!10, dashed, forget plot] coordinates {(0.000001, 0.01) (10000, 100000000)};
            \addplot[color=black!20, dashed, forget plot] coordinates {(0.000001, 0.001) (100000, 100000000)};
            \addplot[color=black!30, dashed, forget plot] coordinates {(0.000001, 0.0001) (1000000, 100000000)};
            \addplot[color=black!40, dashed, forget plot] coordinates {(0.000001, 0.00001) (10000000, 100000000)};
            \addplot[color=black!40, dashed, forget plot] coordinates {(0.00001, 0.000001) (100000000, 10000000)};
            \addplot[color=black!30, dashed, forget plot] coordinates {(0.0001, 0.000001) (100000000, 1000000)};
            \addplot[color=black!20, dashed, forget plot] coordinates {(0.001, 0.000001) (100000000, 100000)};
            \addplot[color=black!10, dashed, forget plot] coordinates {(0.01, 0.000001) (100000000, 10000)};
            \addplot[color=black!5, dashed, forget plot] coordinates {(0.1, 0.000001) (100000000, 1000)};
            \addplot[color=black!3, dashed, forget plot] coordinates {(1, 0.000001) (100000000, 100)};
            \addplot+[only marks, restrict expr to domain={\thisrow{cat}}{0:0}] table[col sep=comma, x=x, y=y]{data/combined-2026-05-07-scatter-sys1-vs-evolved-patterns-time_per_evaluation.csv};
            \addplot+[only marks, restrict expr to domain={\thisrow{cat}}{1:1}] table[col sep=comma, x=x, y=y]{data/combined-2026-05-07-scatter-sys1-vs-evolved-patterns-time_per_evaluation.csv};
            \addplot+[only marks, restrict expr to domain={\thisrow{cat}}{2:2}] table[col sep=comma, x=x, y=y]{data/combined-2026-05-07-scatter-sys1-vs-evolved-patterns-time_per_evaluation.csv};
            \addplot+[only marks, restrict expr to domain={\thisrow{cat}}{3:3}] table[col sep=comma, x=x, y=y]{data/combined-2026-05-07-scatter-sys1-vs-evolved-patterns-time_per_evaluation.csv};
            \addplot+[only marks, restrict expr to domain={\thisrow{cat}}{4:4}] table[col sep=comma, x=x, y=y]{data/combined-2026-05-07-scatter-sys1-vs-evolved-patterns-time_per_evaluation.csv};
            \addplot+[only marks, restrict expr to domain={\thisrow{cat}}{5:5}] table[col sep=comma, x=x, y=y]{data/combined-2026-05-07-scatter-sys1-vs-evolved-patterns-time_per_evaluation.csv};
            \addplot+[only marks, restrict expr to domain={\thisrow{cat}}{6:6}] table[col sep=comma, x=x, y=y]{data/combined-2026-05-07-scatter-sys1-vs-evolved-patterns-time_per_evaluation.csv};
            \addplot+[only marks, restrict expr to domain={\thisrow{cat}}{7:7}] table[col sep=comma, x=x, y=y]{data/combined-2026-05-07-scatter-sys1-vs-evolved-patterns-time_per_evaluation.csv};
            \draw[color=black] (axis cs:1e-70,1e-70) -- (axis cs:1e70,1e70);
      \end{axis}
\end{tikzpicture}%
        \begin{tikzpicture}
      \begin{axis}[
                  xmin=0.000001,
                  xmax=0.001,
                  xtick={0.000001,0.00001,0.0001},
                  extra x ticks={0.001},
                  extra x tick labels={uns.},
                  xmode=log,
                  ymin=0.000001,
                  ymax=0.001,
                  ytick={0.000001,0.00001,0.0001},
                  extra y ticks={0.001},
                  extra y tick labels={uns.},
                  ymode=log,
                  axis equal image,
                  xlabel={\hsystwo{} (lower for 32 tasks)},
                  ylabel={\hevolved{} (lower for 27 tasks)},
                  label style={font=\Large}]
            \addplot[color=black!3, dashed, forget plot] coordinates {(0.000001, 1) (100, 100000000)};
            \addplot[color=black!5, dashed, forget plot] coordinates {(0.000001, 0.1) (1000, 100000000)};
            \addplot[color=black!10, dashed, forget plot] coordinates {(0.000001, 0.01) (10000, 100000000)};
            \addplot[color=black!20, dashed, forget plot] coordinates {(0.000001, 0.001) (100000, 100000000)};
            \addplot[color=black!30, dashed, forget plot] coordinates {(0.000001, 0.0001) (1000000, 100000000)};
            \addplot[color=black!40, dashed, forget plot] coordinates {(0.000001, 0.00001) (10000000, 100000000)};
            \addplot[color=black!40, dashed, forget plot] coordinates {(0.00001, 0.000001) (100000000, 10000000)};
            \addplot[color=black!30, dashed, forget plot] coordinates {(0.0001, 0.000001) (100000000, 1000000)};
            \addplot[color=black!20, dashed, forget plot] coordinates {(0.001, 0.000001) (100000000, 100000)};
            \addplot[color=black!10, dashed, forget plot] coordinates {(0.01, 0.000001) (100000000, 10000)};
            \addplot[color=black!5, dashed, forget plot] coordinates {(0.1, 0.000001) (100000000, 1000)};
            \addplot[color=black!3, dashed, forget plot] coordinates {(1, 0.000001) (100000000, 100)};
            \addplot+[only marks, restrict expr to domain={\thisrow{cat}}{0:0}] table[col sep=comma, x=x, y=y]{data/combined-2026-05-07-scatter-sys2-vs-evolved-patterns-time_per_evaluation.csv};
            \addplot+[only marks, restrict expr to domain={\thisrow{cat}}{1:1}] table[col sep=comma, x=x, y=y]{data/combined-2026-05-07-scatter-sys2-vs-evolved-patterns-time_per_evaluation.csv};
            \addplot+[only marks, restrict expr to domain={\thisrow{cat}}{2:2}] table[col sep=comma, x=x, y=y]{data/combined-2026-05-07-scatter-sys2-vs-evolved-patterns-time_per_evaluation.csv};
            \addplot+[only marks, restrict expr to domain={\thisrow{cat}}{3:3}] table[col sep=comma, x=x, y=y]{data/combined-2026-05-07-scatter-sys2-vs-evolved-patterns-time_per_evaluation.csv};
            \addplot+[only marks, restrict expr to domain={\thisrow{cat}}{4:4}] table[col sep=comma, x=x, y=y]{data/combined-2026-05-07-scatter-sys2-vs-evolved-patterns-time_per_evaluation.csv};
            \addplot+[only marks, restrict expr to domain={\thisrow{cat}}{5:5}] table[col sep=comma, x=x, y=y]{data/combined-2026-05-07-scatter-sys2-vs-evolved-patterns-time_per_evaluation.csv};
            \addplot+[only marks, restrict expr to domain={\thisrow{cat}}{6:6}] table[col sep=comma, x=x, y=y]{data/combined-2026-05-07-scatter-sys2-vs-evolved-patterns-time_per_evaluation.csv};
            \addplot+[only marks, restrict expr to domain={\thisrow{cat}}{7:7}] table[col sep=comma, x=x, y=y]{data/combined-2026-05-07-scatter-sys2-vs-evolved-patterns-time_per_evaluation.csv};
            \draw[color=black] (axis cs:1e-70,1e-70) -- (axis cs:1e70,1e70);
      \end{axis}
\end{tikzpicture}%
    }\\[0.3em]
    \resizebox{\columnwidth}{!}{%
        \begin{tikzpicture}
      \begin{axis}[
                  xmin=0.000001,
                  xmax=0.001,
                  xtick={0.000001,0.00001,0.0001},
                  extra x ticks={0.001},
                  extra x tick labels={uns.},
                  xmode=log,
                  ymin=0.000001,
                  ymax=0.001,
                  ytick={0.000001,0.00001,0.0001},
                  extra y ticks={0.001},
                  extra y tick labels={uns.},
                  ymode=log,
                  axis equal image,
                  xlabel={\hsysthree{} (lower for 19 tasks)},
                  ylabel={\hevolved{} (lower for 56 tasks)},
                  label style={font=\Large}]
            \addplot[color=black!3, dashed, forget plot] coordinates {(0.000001, 1) (100, 100000000)};
            \addplot[color=black!5, dashed, forget plot] coordinates {(0.000001, 0.1) (1000, 100000000)};
            \addplot[color=black!10, dashed, forget plot] coordinates {(0.000001, 0.01) (10000, 100000000)};
            \addplot[color=black!20, dashed, forget plot] coordinates {(0.000001, 0.001) (100000, 100000000)};
            \addplot[color=black!30, dashed, forget plot] coordinates {(0.000001, 0.0001) (1000000, 100000000)};
            \addplot[color=black!40, dashed, forget plot] coordinates {(0.000001, 0.00001) (10000000, 100000000)};
            \addplot[color=black!40, dashed, forget plot] coordinates {(0.00001, 0.000001) (100000000, 10000000)};
            \addplot[color=black!30, dashed, forget plot] coordinates {(0.0001, 0.000001) (100000000, 1000000)};
            \addplot[color=black!20, dashed, forget plot] coordinates {(0.001, 0.000001) (100000000, 100000)};
            \addplot[color=black!10, dashed, forget plot] coordinates {(0.01, 0.000001) (100000000, 10000)};
            \addplot[color=black!5, dashed, forget plot] coordinates {(0.1, 0.000001) (100000000, 1000)};
            \addplot[color=black!3, dashed, forget plot] coordinates {(1, 0.000001) (100000000, 100)};
            \addplot+[only marks, restrict expr to domain={\thisrow{cat}}{0:0}] table[col sep=comma, x=x, y=y]{data/combined-2026-05-07-scatter-sys3-vs-evolved-patterns-time_per_evaluation.csv};
            \addplot+[only marks, restrict expr to domain={\thisrow{cat}}{1:1}] table[col sep=comma, x=x, y=y]{data/combined-2026-05-07-scatter-sys3-vs-evolved-patterns-time_per_evaluation.csv};
            \addplot+[only marks, restrict expr to domain={\thisrow{cat}}{2:2}] table[col sep=comma, x=x, y=y]{data/combined-2026-05-07-scatter-sys3-vs-evolved-patterns-time_per_evaluation.csv};
            \addplot+[only marks, restrict expr to domain={\thisrow{cat}}{3:3}] table[col sep=comma, x=x, y=y]{data/combined-2026-05-07-scatter-sys3-vs-evolved-patterns-time_per_evaluation.csv};
            \addplot+[only marks, restrict expr to domain={\thisrow{cat}}{4:4}] table[col sep=comma, x=x, y=y]{data/combined-2026-05-07-scatter-sys3-vs-evolved-patterns-time_per_evaluation.csv};
            \addplot+[only marks, restrict expr to domain={\thisrow{cat}}{5:5}] table[col sep=comma, x=x, y=y]{data/combined-2026-05-07-scatter-sys3-vs-evolved-patterns-time_per_evaluation.csv};
            \addplot+[only marks, restrict expr to domain={\thisrow{cat}}{6:6}] table[col sep=comma, x=x, y=y]{data/combined-2026-05-07-scatter-sys3-vs-evolved-patterns-time_per_evaluation.csv};
            \addplot+[only marks, restrict expr to domain={\thisrow{cat}}{7:7}] table[col sep=comma, x=x, y=y]{data/combined-2026-05-07-scatter-sys3-vs-evolved-patterns-time_per_evaluation.csv};
            \draw[color=black] (axis cs:1e-70,1e-70) -- (axis cs:1e70,1e70);
      \end{axis}
\end{tikzpicture}%
        \begin{tikzpicture}
      \begin{axis}[
                  xmin=0.000001,
                  xmax=0.001,
                  xtick={0.000001,0.00001,0.0001},
                  extra x ticks={0.001},
                  extra x tick labels={uns.},
                  xmode=log,
                  ymin=0.000001,
                  ymax=0.001,
                  ytick={0.000001,0.00001,0.0001},
                  extra y ticks={0.001},
                  extra y tick labels={uns.},
                  ymode=log,
                  axis equal image,
                  xlabel={\hsysscp{} (lower for 25 tasks)},
                  ylabel={\hevolved{} (lower for 50 tasks)},
                  label style={font=\Large}]
            \addplot[color=black!3, dashed, forget plot] coordinates {(0.000001, 1) (100, 100000000)};
            \addplot[color=black!5, dashed, forget plot] coordinates {(0.000001, 0.1) (1000, 100000000)};
            \addplot[color=black!10, dashed, forget plot] coordinates {(0.000001, 0.01) (10000, 100000000)};
            \addplot[color=black!20, dashed, forget plot] coordinates {(0.000001, 0.001) (100000, 100000000)};
            \addplot[color=black!30, dashed, forget plot] coordinates {(0.000001, 0.0001) (1000000, 100000000)};
            \addplot[color=black!40, dashed, forget plot] coordinates {(0.000001, 0.00001) (10000000, 100000000)};
            \addplot[color=black!40, dashed, forget plot] coordinates {(0.00001, 0.000001) (100000000, 10000000)};
            \addplot[color=black!30, dashed, forget plot] coordinates {(0.0001, 0.000001) (100000000, 1000000)};
            \addplot[color=black!20, dashed, forget plot] coordinates {(0.001, 0.000001) (100000000, 100000)};
            \addplot[color=black!10, dashed, forget plot] coordinates {(0.01, 0.000001) (100000000, 10000)};
            \addplot[color=black!5, dashed, forget plot] coordinates {(0.1, 0.000001) (100000000, 1000)};
            \addplot[color=black!3, dashed, forget plot] coordinates {(1, 0.000001) (100000000, 100)};
            \addplot+[only marks, restrict expr to domain={\thisrow{cat}}{0:0}] table[col sep=comma, x=x, y=y]{data/combined-2026-05-07-scatter-sys-scp-vs-evolved-patterns-time_per_evaluation.csv};
            \addplot+[only marks, restrict expr to domain={\thisrow{cat}}{1:1}] table[col sep=comma, x=x, y=y]{data/combined-2026-05-07-scatter-sys-scp-vs-evolved-patterns-time_per_evaluation.csv};
            \addplot+[only marks, restrict expr to domain={\thisrow{cat}}{2:2}] table[col sep=comma, x=x, y=y]{data/combined-2026-05-07-scatter-sys-scp-vs-evolved-patterns-time_per_evaluation.csv};
            \addplot+[only marks, restrict expr to domain={\thisrow{cat}}{3:3}] table[col sep=comma, x=x, y=y]{data/combined-2026-05-07-scatter-sys-scp-vs-evolved-patterns-time_per_evaluation.csv};
            \addplot+[only marks, restrict expr to domain={\thisrow{cat}}{4:4}] table[col sep=comma, x=x, y=y]{data/combined-2026-05-07-scatter-sys-scp-vs-evolved-patterns-time_per_evaluation.csv};
            \addplot+[only marks, restrict expr to domain={\thisrow{cat}}{5:5}] table[col sep=comma, x=x, y=y]{data/combined-2026-05-07-scatter-sys-scp-vs-evolved-patterns-time_per_evaluation.csv};
            \addplot+[only marks, restrict expr to domain={\thisrow{cat}}{6:6}] table[col sep=comma, x=x, y=y]{data/combined-2026-05-07-scatter-sys-scp-vs-evolved-patterns-time_per_evaluation.csv};
            \addplot+[only marks, restrict expr to domain={\thisrow{cat}}{7:7}] table[col sep=comma, x=x, y=y]{data/combined-2026-05-07-scatter-sys-scp-vs-evolved-patterns-time_per_evaluation.csv};
            \draw[color=black] (axis cs:1e-70,1e-70) -- (axis cs:1e70,1e70);
      \end{axis}
\end{tikzpicture}%
    }%
    \par\vspace{0.5em}%
    {\centering\pgfplotslegendfromname{scatterlegend}\par}%
    \caption{Time per heuristic evaluation: \hevolved{} vs.\ each baseline. Points
        below the diagonal indicate tasks where \hevolved{} evaluates each state
        more cheaply.}
    \label{plot:time_per_eval}
\end{figure}

Figure~\ref{plot:time_per_eval} isolates per-state evaluation cost and reveals the mechanism behind the total-time advantage.
\hevolved{} is faster than \hsysthree{} on 56 tasks out of 75, and faster than \hsysscp{} on 50 tasks out of 75.
Although the patterns in \hevolved{} are individually larger, the collection generally contains fewer patterns, which reduces the number of lookups per state evaluation.
Across the 40 tasks solved by all five approaches, \hevolved{} produces 1230 patterns in total, compared to 5998 for \hsysscp{}, 49\,157 for \hsysthree{}, 2028 for \hsystwo{} and 267 for \hsysone{}.
This per-state efficiency is the primary reason \hevolved{} matches or exceeds stronger heuristics in most domains, despite being less informative and requiring more expansions on average, and it is precisely what gives our patterns an edge over the strongest systematic baselines.
The exception is again Miconic, where the per-state evaluation cost of \hsysthree{} and \hsysscp{} is more than offset by the information their heuristics provide.

\subsection{Example Pattern Collection Generators}
In this section, we describe pattern collection generators synthesized for three representative domains and analyze their empirical behavior in more depth.

\subparagraph{Childsnack.}
In Childsnack, sandwiches must be assembled and delivered to children waiting at tables, some of whom require gluten-free sandwiches.
The pattern generator found by our approach produces seven pattern types: \emph{goal singletons} (1 atom); \emph{tray location mutexes} (2--4 atoms); \emph{kitchen resource pools} grouping all ingredients in the kitchen (2--20 atoms); \emph{sandwich life-cycle patterns} tracking one sandwich through its states together with its gluten status (5--13 atoms); \emph{gluten-allergy constraints} aggregating served-goals of allergic children with gluten-free sandwich atoms (2--20 atoms); \emph{place-based delivery patterns} collecting all trays, sandwiches and served-goals at one delivery location (2--49 atoms); and \emph{goal-delivery context patterns} pairing one child's served-goal with the trays, sandwiches and gluten constraints at that child's location (4--64 atoms).

These patterns are substantially larger than those found by systematic methods: \hevolved{} produces patterns of up to size 64, while \hsysscp{} stops at size 2.
On every solved task, \hevolved{} therefore needs fewer expansions before the last $f$-layer than any systematic baseline---a gap of roughly one order of magnitude---and it achieves the highest coverage on the domain (5 vs.\ 4 for all baselines).

\subparagraph{Miconic.}
Miconic is an elevator scheduling domain where a single lift must board and deliver passengers to their destination floors.
Our generator produces three pattern types: \emph{singleton state atoms} (\texttt{boarded} or \texttt{served}, 1 atom each); \emph{lift floor patterns} grouping \texttt{lift-at} atoms for adjacent or globally relevant floor pairs (1--19 atoms); and \emph{per-passenger journey patterns} capturing the full life cycle of one passenger together with the lift positions at the passenger's origin and destination, optionally extended to passengers sharing a floor (5--19 atoms).

Miconic is the domain where our approach struggles the most.
Despite using patterns of up to size 8, \hevolved{} solves only 4 of 30 tasks, compared to 20 for \hsysthree{}.
Figure~\ref{plot:expansions} confirms that \hsysthree{} and \hsysscp{} need far fewer expansions on nearly every solved task (16--20 out of 20), so small systematic patterns provide stronger guidance here.
Miconic's structure makes the domain particularly amenable to exhaustive pattern enumeration: its uniform passenger--floor structure means that patterns of size 3 already capture the key interactions, leaving little room for the larger but less comprehensive patterns produced by our generator.

\subparagraph{Transport.}
Transport is a logistics domain where vehicles with limited capacity move packages between locations on a road network.
Our pattern generator produces four pattern types: \emph{per-package delivery patterns} combining a package's initial and goal locations with all possible in-vehicle states (3--20 atoms); \emph{per-vehicle cargo patterns} collecting all packages loadable into one vehicle together with capacity atoms (3--20 atoms); \emph{per-vehicle movement patterns} grouping all locations a vehicle can occupy with its capacity levels (4--20 atoms); and \emph{location clusters} grouping all objects at initial or goal locations of any package, with uncovered goal atoms added as singletons (1--20 atoms).

Transport is one of the strongest domains for \hevolved{}, which achieves the best coverage (13 solved tasks vs.\ 12 for \hsysthree{}).
Guidance quality, measured in expansions before the last $f$-layer, is comparable to \hsystwo{} and \hsysthree{} on solved tasks.
The main advantage of our patterns is their lower evaluation cost: \hevolved{} is faster to evaluate than \hsysthree{} on 11 tasks out of 12, and faster than \hsysscp{} on 10 tasks out of 12.
This per-state efficiency translates directly into lower total search time and higher coverage on hard instances.

\section{Related Work}
\label{sec:related}
\paragraph{Pattern Database Heuristics.}
Pattern databases (PDBs) \cite{culberson-schaeffer-compint1998} are one of several admissible heuristic families for optimal classical planning, alongside merge-and-shrink abstractions \cite{helmert-et-al-jacm2014,sievers-helmert-jair2021}, Cartesian abstractions \cite{seipp-helmert-jair2018}, LP-based operator-counting heuristics \cite{pommerening-et-al-icaps2014} and landmark-based heuristics such as \mbox{LM-cut} \cite{helmert-domshlak-icaps2009}. Domain-independent generators for PDBs include the hill-climbing procedure of \citet{haslum-et-al-aaai2007} and counterexample-guided pattern selection \cite{rovner-et-al-icaps2019}. \citet{franco-et-al-ijcai2017} construct \emph{complementary} pattern collections iteratively: at each step they propose a new collection with a bin-packing-based generator and accept it only if a sampling-based estimator predicts that adding it reduces \astar{} search effort, biasing new patterns toward states the current heuristic underestimates. Systematic pattern generation \cite{pommerening-et-al-ijcai2013} identifies \emph{interesting} patterns by exploiting causal-graph
structure, and \citet{seipp-ijcai2019} extends this approach to the current state of the art by only selecting those
systematic patterns that saturated cost partitioning deems useful. Closest in spirit to our work,
\citet{edelkamp-mochart2006} uses a genetic algorithm to evolve pattern collections directly, but
the search operates over a fixed encoding of patterns rather than synthesizing a generator program.
All these methods are domain-independent and treat each task in isolation, so any structure shared across tasks of the same domain has to be rediscovered every time. We lift this restriction by learning, per domain, a generator whose code captures regularities the LLM distills from a handful of training tasks.

\paragraph{Learning Heuristics for Planning.}
A growing line of work learns heuristics for satisficing planning from experience \egcite{toyer-et-al-aaai2018,stahlberg-et-al-icaps2022}, with \inlinecite{chen-et-al-aaai2024} extending this to heuristics that transfer across domains. \inlinecite{chen-et-al-icaps2024} further show that classical machine-learning models can match neural heuristics at substantially lower inference cost, and \inlinecite{karia-srivastava-aaai2021} learn relational heuristics that transfer across object counts. \inlinecite{frances-et-al-ijcai2019} take a symbolic route, synthesizing per-domain generalized potential heuristics over a fixed concept-feature language. These heuristics are typically inadmissible, and the neural variants often require a GPU at inference time. \inlinecite{nunez-molina-et-al-ijcai2024} bound learned values by admissible estimates during training to mitigate the second issue, but their final heuristic remains inadmissible. Admissible learned heuristics are rarer. Closest in spirit, \citet{futuhi-sturtevant-iclr2026} train a general-purpose neural admissible heuristic with a cross-entropy admissibility loss, but only enforce admissibility on the training distribution rather than by construction.

\paragraph{Evolutionary and LLM-Guided Heuristic Search.}
Automatic heuristic discovery predates LLMs. \inlinecite{aler-et-al-evco2001} use genetic programming over a fixed language to evolve domain-specific control heuristics, and \inlinecite{fukunaga-evco2008} evolves composite SAT heuristics from a fixed compositional grammar. Since FunSearch \cite{romera-paredes-et-al-nature2024}, the field has moved away from such restricted languages and uses LLMs to mutate raw code instead, with strong results across diverse task families \egcite{liu-et-al-icml2024,ye-et-al-neurips2024,zheng-et-al-icml2025,novikov-et-al-arxiv2025}. We follow this trend and build on the open-source OpenEvolve framework \cite{sharma-misc2025}. To our knowledge, ours is the first application of LLM-driven program synthesis to admissible heuristics for optimal planning.

\paragraph{LLM-Generated Domain-Specific Solvers.}
Several recent methods use LLMs to produce domain-specific solvers. \inlinecite{correa-et-al-neurips2025} and \inlinecite{tuisov-et-al-icaps2026} generate inadmissible heuristics that let otherwise weak planners compete with strong ones. Our generators are domain-specific in the same sense, but the evolutionary process is identical across domains and the generators produce \emph{admissible} heuristics that plug directly into a state-of-the-art planner. A separate strand generates domain-specific policies \egcite{chen-et-al-icaps2025wslm4plan,stein-et-al-icaps2026}. Closest in spirit, \inlinecite{murray-et-al-icaps2026} use LLM-driven evolution to produce Python functions that emit plans directly, a form of generalized planning competitive with state-of-the-art planners on their evaluation set. These policy-style approaches and ours are complementary: a learned policy can solve target tasks quickly and often without search but cannot scale beyond tasks that admit simple strategies, whereas a learned heuristic preserves the search guarantees that make symbolic planners broadly applicable.

\section{Conclusions}
We presented a method for learning admissible heuristics for classical planning by evolving domain-specific pattern collection generators with an LLM-guided evolutionary framework. The evolved generators are interpretable Python programs that compute a pattern collection for any task in the domain at test time and yield admissible heuristics by construction. Across seven domains, the evolved heuristic matches or exceeds the best systematic baseline in five, and is uniquely best in two of them. On a majority of tasks it is also substantially faster to evaluate per state than the strongest systematic baselines, often by orders of magnitude, thanks to the task-focused structure of the evolved patterns. The two domains where our approach falls behind, Floortile and Miconic, are precisely the ones in which small systematic patterns already provide strong guidance and leave little room for the larger domain-specific ones our generators compute.

Several directions stand out for future work. First, exposing the evolutionary algorithm to richer domain information, such as the causal graph or the domain transition graph, may improve pattern quality. Second, we trained on a small set of tasks per domain to keep token usage manageable, and larger training sets may uncover stronger generators. Third, the slight mismatch between our training score and the final evaluation metric (coverage) suggests that refining the scoring function could yield further gains. Finally, obfuscating domain names would clarify how much our approach relies on structural reasoning rather than on the LLM's prior exposure to these benchmarks.

\section*{Acknowledgments}

This work is supported by the Swedish Research Council under grant number 2024-05403, and by the Wallenberg
AI, Autonomous Systems and Software Program (WASP)
funded by the Knut and Alice Wallenberg Foundation.
The computations were enabled by resources
provided by the National Academic Infrastructure for
Supercomputing in Sweden (NAISS), partially funded by
the Swedish Research Council.

\bibliography{abbrv,literatur,crossref,extra}

\onecolumn

\appendix

\section{Synthesized Pattern Generators}

\captionof{lstlisting}{Blocksworld generator code.}
\begin{lstlisting}[
    language=python,
    backgroundcolor=\color{backcolour},
    keywordstyle=\color{violet},
    basicstyle=\ttfamily\footnotesize,
    breakatwhitespace=false,
    breaklines=true,
    keepspaces=true,
    showspaces=false,
    numbers=none,
    showstringspaces=false,
    xleftmargin=0pt,
    xrightmargin=0pt,
]

def generate_pattern_collection(task_info: TaskInformation) -> list[Pattern]:
    patterns = []
    goal_atoms = list(task_info.fluent_goal_atoms)
    
    if not goal_atoms:
        if task_info.fluent_initial_state_atoms:
            patterns.append(Pattern(pattern=[task_info.fluent_initial_state_atoms[0]]))
        return patterns
    
    obj_to_atoms = {}
    for atom in goal_atoms:
        for obj in atom.binding:
            if obj not in obj_to_atoms:
                obj_to_atoms[obj] = []
            obj_to_atoms[obj].append(atom)
    
    visited = set()
    
    def dfs(atom, component):
        atom_id = id(atom)
        if atom_id in visited:
            return
        visited.add(atom_id)
        component.append(atom)
        for obj in atom.binding:
            for other in obj_to_atoms.get(obj, []):
                dfs(other, component)
    
    for atom in goal_atoms:
        if id(atom) not in visited:
            component = []
            dfs(atom, component)
            if len(component) > 1:
                patterns.append(Pattern(pattern=component))
    
    clear_map = {}
    holding_map = {}
    ontable_map = {}
    on_map = {}
    arm_empty_atom = None
    
    for atom in task_info.all_fluent_atoms:
        pname = atom.predicate.name
        if pname == 'clear' and len(atom.binding) == 1:
            clear_map[atom.binding[0]] = atom
        elif pname == 'holding' and len(atom.binding) == 1:
            holding_map[atom.binding[0]] = atom
        elif pname == 'on-table' and len(atom.binding) == 1:
            ontable_map[atom.binding[0]] = atom
        elif pname == 'on' and len(atom.binding) == 2:
            on_map[atom.binding] = atom
        elif pname == 'arm-empty':
            arm_empty_atom = atom
    
    goal_objs = set()
    for atom in goal_atoms:
        for obj in atom.binding:
            goal_objs.add(obj)
    
    for atom in goal_atoms:
        pattern_atoms = [atom]
        objs = set(atom.binding)
        
        for fluent_atom in task_info.all_fluent_atoms:
            if len(pattern_atoms) >= 6:
                break
            if fluent_atom.predicate.name == 'clear':
                if fluent_atom.binding[0] in objs:
                    pattern_atoms.append(fluent_atom)
        
        if arm_empty_atom and len(pattern_atoms) < 8:
            pattern_atoms.append(arm_empty_atom)
        
        patterns.append(Pattern(pattern=pattern_atoms))
    
    for obj in goal_objs:
        if obj in holding_map:
            patterns.append(Pattern(pattern=[holding_map[obj]]))
    
    initial_on = {a for a in task_info.fluent_initial_state_atoms if a.predicate.name == 'on'}
    goal_on = {a for a in goal_atoms if a.predicate.name == 'on'}
    initial_support = {a.binding[0]: a.binding[1] for a in initial_on if len(a.binding) == 2}
    goal_support = {a.binding[0]: a.binding[1] for a in goal_on if len(a.binding) == 2}
    
    blocks_to_unstack = set()
    blocks_to_pickup = set()
    blocks_to_putdown = set()
    
    for block in goal_objs:
        if block in initial_support:
            if block not in goal_support or initial_support[block] != goal_support[block]:
                blocks_to_unstack.add(block)
    
    for atom in goal_atoms:
        if atom.predicate.name == 'on' and len(atom.binding) == 2:
            block = atom.binding[0]
            if block in ontable_map and block in clear_map:
                blocks_to_pickup.add(block)
    
    for atom in goal_atoms:
        if atom.predicate.name == 'on-table' and len(atom.binding) == 1:
            block = atom.binding[0]
            if block in holding_map:
                blocks_to_putdown.add(block)
    
    for atom in goal_atoms:
        if atom.predicate.name == 'on' and len(atom.binding) == 2:
            x, y = atom.binding[0], atom.binding[1]
            if x in holding_map and y in clear_map:
                stack_pattern = [atom, holding_map[x], clear_map[y]]
                if arm_empty_atom:
                    stack_pattern.append(arm_empty_atom)
                if len(stack_pattern) <= 15:
                    patterns.append(Pattern(pattern=stack_pattern))
    
    for block in blocks_to_unstack:
        if block in initial_support and block in clear_map and arm_empty_atom:
            y = initial_support[block]
            if (block, y) in on_map:
                unstack_pattern = [on_map[(block, y)], clear_map[block], arm_empty_atom]
                if len(unstack_pattern) <= 12:
                    patterns.append(Pattern(pattern=unstack_pattern))
    
    for block in blocks_to_pickup:
        if block in ontable_map and block in clear_map and arm_empty_atom:
            pickup_pattern = [ontable_map[block], clear_map[block], arm_empty_atom]
            if len(pickup_pattern) <= 10:
                patterns.append(Pattern(pattern=pickup_pattern))
    
    for block in blocks_to_putdown:
        if block in holding_map and block in ontable_map:
            putdown_pattern = [ontable_map[block], holding_map[block]]
            if len(putdown_pattern) <= 10:
                patterns.append(Pattern(pattern=putdown_pattern))
    
    initial_above = {}
    for a in initial_on:
        if len(a.binding) == 2:
            initial_above[a.binding[1]] = a.binding[0]
    
    for atom in goal_atoms:
        if atom.predicate.name == 'on' and len(atom.binding) == 2:
            y = atom.binding[1]
            if y in initial_above:
                z = initial_above[y]
                if z in clear_map and arm_empty_atom and (z, y) in on_map:
                    obstacle_pattern = [atom, on_map[(z, y)], clear_map[z], arm_empty_atom]
                    if len(obstacle_pattern) <= 12:
                        patterns.append(Pattern(pattern=obstacle_pattern))
    
    for atom in goal_atoms:
        patterns.append(Pattern(pattern=[atom]))
    
    for atom in goal_atoms:
        patterns.append(Pattern(pattern=[atom]))
    
    seen = set()
    unique_patterns = []
    for p in patterns:
        key = tuple(sorted(id(atom) for atom in p.pattern))
        if key not in seen and len(p.pattern) <= 20:
            seen.add(key)
            unique_patterns.append(p)
    
    goal_atom_set = set(goal_atoms)
    
    def pattern_value(p):
        coverage = len(set(p.pattern) & goal_atom_set)
        size = len(p.pattern)
        cost = 2 ** size
        
        if coverage == 0:
            return -cost * 0.1 if size <= 4 else -cost
        
        value = (coverage * 1000.0) / cost
        if size >= 12:
            value *= 0.1
        elif size >= 8:
            value *= 0.4
        elif size >= 6:
            value *= 0.7
        
        return value
    
    unique_patterns.sort(key=pattern_value, reverse=True)
    
    selected_patterns = []
    covered_goals = set()
    total_states = 0
    
    for p in unique_patterns:
        cost = 2 ** len(p.pattern)
        if total_states + cost <= 500000000:
            selected_patterns.append(p)
            total_states += cost
            covered_goals.update(set(p.pattern) & goal_atom_set)
    
    for goal in goal_atom_set - covered_goals:
        if total_states + 2 <= 500000000:
            selected_patterns.append(Pattern(pattern=[goal]))
            total_states += 2
    
    return selected_patterns
\end{lstlisting}

\captionof{lstlisting}{Childsnack generator code.}
\begin{lstlisting}[
    language=python,
    backgroundcolor=\color{backcolour},
    keywordstyle=\color{violet},
    basicstyle=\ttfamily\footnotesize,
    breakatwhitespace=false,
    breaklines=true,
    keepspaces=true,
    showspaces=false,
    numbers=none,
    showstringspaces=false,
    xleftmargin=0pt,
    xrightmargin=0pt,
]

def generate_pattern_collection(task_info: TaskInformation) -> list[Pattern]:
    patterns = []
    
    def count_vars(atom_list):
        return len(set(o for atom in atom_list for o in atom.binding))
    
    fluent_by_pred = {}
    for atom in task_info.all_fluent_atoms:
        fluent_by_pred.setdefault(atom.predicate.name, []).append(atom)
    
    static_by_pred = {}
    for atom in task_info.static_ground_atoms:
        static_by_pred.setdefault(atom.predicate.name, []).append(atom)
    
    for goal in task_info.fluent_goal_atoms:
        patterns.append(Pattern(pattern=[goal]))
    
    at_atoms = fluent_by_pred.get('at', [])
    trays = {}
    for atom in at_atoms:
        trays.setdefault(atom.binding[0], []).append(atom)
    
    for tray_atoms in trays.values():
        if count_vars(tray_atoms) <= 20:
            patterns.append(Pattern(pattern=tray_atoms))
    
    notexist = {a.binding[0]: a for a in fluent_by_pred.get('notexist', [])}
    at_kitchen = {a.binding[0]: a for a in fluent_by_pred.get('at_kitchen_sandwich', [])}
    no_gluten = {a.binding[0]: a for a in fluent_by_pred.get('no_gluten_sandwich', [])}
    ontray_by_sw = {}
    for atom in fluent_by_pred.get('ontray', []):
        ontray_by_sw.setdefault(atom.binding[0], []).append(atom)
    
    tray_to_loc = {}
    for atom in at_atoms:
        tray_to_loc[atom.binding[0]] = atom
    
    for sw in set(notexist.keys()) | set(at_kitchen.keys()) | set(ontray_by_sw.keys()):
        atoms = []
        if sw in notexist: atoms.append(notexist[sw])
        if sw in at_kitchen: atoms.append(at_kitchen[sw])
        if sw in ontray_by_sw: 
            atoms.extend(ontray_by_sw[sw])
            for ontray_atom in ontray_by_sw[sw]:
                tray = ontray_atom.binding[1]
                if tray in tray_to_loc:
                    atoms.append(tray_to_loc[tray])
        if sw in no_gluten: atoms.append(no_gluten[sw])
        if 1 < count_vars(atoms) <= 20:
            patterns.append(Pattern(pattern=atoms))
    
    bread = fluent_by_pred.get('at_kitchen_bread', [])
    content = fluent_by_pred.get('at_kitchen_content', [])
    if bread or content:
        atoms = bread + content
        if 1 < count_vars(atoms) <= 20:
            patterns.append(Pattern(pattern=atoms))
    
    waiting = {a.binding[0]: a.binding[1] for a in static_by_pred.get('waiting', [])}
    allergic = {a.binding[0] for a in static_by_pred.get('allergic_gluten', [])}
    no_gluten_set = set(no_gluten.keys())
    ontray_atoms = fluent_by_pred.get('ontray', [])
    
    for goal in task_info.fluent_goal_atoms:
        if goal.predicate.name != 'served':
            continue
        child = goal.binding[0]
        if child not in waiting:
            continue
            
        place = waiting[child]
        atoms = [goal]
        
        trays_here = []
        for atom in at_atoms:
            if atom.binding[1] == place:
                atoms.append(atom)
                trays_here.append(atom.binding[0])
        
        valid_sandwiches = set()
        for atom in ontray_atoms:
            if atom.binding[1] in trays_here:
                if child not in allergic or atom.binding[0] in no_gluten_set:
                    atoms.append(atom)
                    valid_sandwiches.add(atom.binding[0])
        
        if child in allergic:
            for sw_id in valid_sandwiches:
                if sw_id in no_gluten:
                    atoms.append(no_gluten[sw_id])
        
        if 1 < count_vars(atoms) <= 20:
            patterns.append(Pattern(pattern=atoms))
    
    if allergic:
        atoms = [g for g in task_info.fluent_goal_atoms 
                if g.predicate.name == 'served' and g.binding[0] in allergic]
        for atom in fluent_by_pred.get('no_gluten_sandwich', []):
            if count_vars(atoms + [atom]) <= 20:
                atoms.append(atom)
        if len(atoms) > 1 and count_vars(atoms) <= 20:
            patterns.append(Pattern(pattern=atoms))
    
    places = set()
    for atom in at_atoms:
        places.add(atom.binding[1])
    
    for place in places:
        place_atoms = []
        trays_here = []
        
        for atom in at_atoms:
            if atom.binding[1] == place:
                place_atoms.append(atom)
                trays_here.append(atom.binding[0])
        
        for atom in ontray_atoms:
            if atom.binding[1] in trays_here:
                place_atoms.append(atom)
        
        for goal in task_info.fluent_goal_atoms:
            if goal.predicate.name == 'served':
                child = goal.binding[0]
                if child in waiting and waiting[child] == place:
                    place_atoms.append(goal)
        
        if 1 < count_vars(place_atoms) <= 20:
            patterns.append(Pattern(pattern=place_atoms))
    
    seen = set()
    unique = []
    for p in patterns:
        key = frozenset((a.predicate.name, a.binding) for a in p.pattern)
        if key not in seen:
            seen.add(key)
            unique.append(p)
    
    return unique
\end{lstlisting}

\captionof{lstlisting}{Floortile generator code.}
\begin{lstlisting}[
    language=python,
    backgroundcolor=\color{backcolour},
    keywordstyle=\color{violet},
    basicstyle=\ttfamily\footnotesize,
    breakatwhitespace=false,
    breaklines=true,
    keepspaces=true,
    showspaces=false,
    numbers=none,
    showstringspaces=false,
    xleftmargin=0pt,
    xrightmargin=0pt,
]

def generate_pattern_collection(task_info: TaskInformation) -> list[Pattern]:
    patterns = []
    
    def get_predicate_name(atom):
        return atom.predicate.name
    
    paint_adjacency = {}
    move_adjacency = {}
    
    for atom in task_info.static_ground_atoms:
        pname = get_predicate_name(atom)
        if pname == 'up':
            target, source = atom.binding  # target is up from source
            if target not in paint_adjacency:
                paint_adjacency[target] = []
            paint_adjacency[target].append(source)
            if source not in move_adjacency:
                move_adjacency[source] = []
            move_adjacency[source].append(target)
            if target not in move_adjacency:
                move_adjacency[target] = []
            move_adjacency[target].append(source)
        elif pname == 'down':
            target, source = atom.binding  # target is down from source
            if target not in paint_adjacency:
                paint_adjacency[target] = []
            paint_adjacency[target].append(source)
            if source not in move_adjacency:
                move_adjacency[source] = []
            move_adjacency[source].append(target)
            if target not in move_adjacency:
                move_adjacency[target] = []
            move_adjacency[target].append(source)
        elif pname == 'left':
            target, source = atom.binding
            if source not in move_adjacency:
                move_adjacency[source] = []
            move_adjacency[source].append(target)
            if target not in move_adjacency:
                move_adjacency[target] = []
            move_adjacency[target].append(source)
        elif pname == 'right':
            target, source = atom.binding
            if source not in move_adjacency:
                move_adjacency[source] = []
            move_adjacency[source].append(target)
            if target not in move_adjacency:
                move_adjacency[target] = []
            move_adjacency[target].append(source)
    
    goal_tiles = {}  # tile -> color
    robot_at_atoms = []
    robot_has_atoms = []
    painted_atoms = []
    clear_atoms = []
    tile_to_clear_atom = {}
    
    for atom in task_info.fluent_goal_atoms:
        if get_predicate_name(atom) == 'painted':
            tile, color = atom.binding
            goal_tiles[tile] = color
    
    for atom in task_info.all_fluent_atoms:
        pname = get_predicate_name(atom)
        if pname == 'robot-at':
            robot_at_atoms.append(atom)
        elif pname == 'robot-has':
            robot_has_atoms.append(atom)
        elif pname == 'painted':
            painted_atoms.append(atom)
        elif pname == 'clear':
            clear_atoms.append(atom)
            tile_to_clear_atom[atom.binding[0]] = atom
    
    init_robot_at = None
    for atom in task_info.fluent_initial_state_atoms:
        if get_predicate_name(atom) == 'robot-at':
            init_robot_at = atom
            break
    
    goal_tiles_set = set(goal_tiles.keys())
    
    for goal_atom in task_info.fluent_goal_atoms:
        if get_predicate_name(goal_atom) != 'painted':
            continue
            
        pattern_atoms = [goal_atom]
        tile, color = goal_atom.binding
        
        if init_robot_at is not None:
            patterns.append(Pattern(pattern=[goal_atom, init_robot_at]))
        
        for rha in robot_has_atoms:
            if rha.binding[1] == color:
                pattern_atoms.append(rha)
                break
        
        if tile in paint_adjacency:
            for adj_tile in paint_adjacency[tile]:
                for raa in robot_at_atoms:
                    if raa.binding[1] == adj_tile:
                        pattern_atoms.append(raa)
                        break
        
        if tile in tile_to_clear_atom:
            pattern_atoms.append(tile_to_clear_atom[tile])
        if tile in paint_adjacency:
            for adj_tile in paint_adjacency[tile]:
                if adj_tile in tile_to_clear_atom:
                    pattern_atoms.append(tile_to_clear_atom[adj_tile])
        
        unique_atoms = []
        seen = set()
        for a in pattern_atoms:
            key = (a.predicate.name, tuple(o.name for o in a.binding))
            if key not in seen:
                seen.add(key)
                unique_atoms.append(a)
        
        if len(unique_atoms) <= 20:
            patterns.append(Pattern(pattern=unique_atoms))
        else:
            patterns.append(Pattern(pattern=[goal_atom]))
    
    robot_at_by_tile = {}
    for ra in robot_at_atoms:
        robot_at_by_tile.setdefault(ra.binding[1], []).append(ra)
    
    for atom in task_info.static_ground_atoms:
        if atom.predicate.name not in ['up', 'down']:
            continue
        target, source = atom.binding[0], atom.binding[1]
        if target not in goal_tiles_set:
            continue
        for ratom in robot_at_by_tile.get(source, []):
            robot = ratom.binding[0]
            for catom in clear_atoms:
                if catom.binding[0] == target:
                    for hasatom in robot_has_atoms:
                        if hasatom.binding[0] == robot:
                            precond_pattern = [ratom, hasatom, catom]
                            if len(precond_pattern) <= 15:
                                patterns.append(Pattern(pattern=precond_pattern))
    
    if robot_at_atoms and robot_has_atoms:
        robot_pattern = list(dict.fromkeys(robot_at_atoms + robot_has_atoms))
        relevant_clears = set()
        for tile in goal_tiles:
            if tile in paint_adjacency:
                for adj_tile in paint_adjacency[tile]:
                    if adj_tile in tile_to_clear_atom:
                        relevant_clears.add(tile_to_clear_atom[adj_tile])
        robot_pattern.extend(list(relevant_clears)[:15])
        robot_pattern = list(dict.fromkeys(robot_pattern))
        if len(robot_pattern) <= 20:
            patterns.append(Pattern(pattern=robot_pattern))
    
    clear_by_tile = {a.binding[0]: a for a in clear_atoms}
    tile_groups = {}
    
    for tile in clear_by_tile:
        parts = tile.name.split('_')
        if len(parts) >= 3:
            row, col = parts[1], parts[2]
            tile_groups.setdefault(('r', row), []).append(clear_by_tile[tile])
            tile_groups.setdefault(('c', col), []).append(clear_by_tile[tile])
    
    for group in tile_groups.values():
        if 2 <= len(group) <= 8:
            patterns.append(Pattern(pattern=group))
    
    robots = set()
    for atom in robot_at_atoms:
        robots.add(atom.binding[0])
    for robot in robots:
        has_atoms = [a for a in robot_has_atoms if a.binding[0] == robot]
        if 1 < len(has_atoms) <= 20:
            patterns.append(Pattern(pattern=has_atoms))
    
    all_tiles = set()
    for atom in clear_atoms + painted_atoms:
        all_tiles.add(atom.binding[0])
    for tile in all_tiles:
        tile_atoms = ([a for a in clear_atoms if a.binding[0] == tile] +
                     [a for a in painted_atoms if a.binding[0] == tile])
        if 1 < len(tile_atoms) <= 20:
            patterns.append(Pattern(pattern=tile_atoms))
    
    color_to_goals = {}
    for tile, color in goal_tiles.items():
        if color not in color_to_goals:
            color_to_goals[color] = []
        for pa in painted_atoms:
            if pa.binding[0] == tile and pa.binding[1] == color:
                color_to_goals[color].append(pa)
                break
    
    for color, goal_atoms in color_to_goals.items():
        if goal_atoms:
            color_pattern = list(dict.fromkeys(goal_atoms + robot_has_atoms))
            color_pattern.extend(robot_at_atoms[:2])
            if len(color_pattern) <= 20:
                patterns.append(Pattern(pattern=color_pattern))
    
    seen = set()
    unique_patterns = []
    for p in patterns:
        key = tuple(sorted([(a.predicate.name, tuple(o.name for o in a.binding)) for a in p.pattern]))
        if key not in seen and len(p.pattern) > 0:
            seen.add(key)
            unique_patterns.append(p)
    
    def pattern_priority(p):
        covers_goal = any(atom in task_info.fluent_goal_atoms for atom in p.pattern)
        return (0 if covers_goal else 1, len(p.pattern))
    
    unique_patterns.sort(key=pattern_priority)
    
    selected = []
    total_states = 0
    budget = 400000000  # 400M limit (under 500M)
    
    for p in unique_patterns:
        n_atoms = len(p.pattern)
        if n_atoms > 20:  # Enforce per-pattern atom limit
            continue
        states = 2 ** n_atoms
        if total_states + states <= budget:
            selected.append(p)
            total_states += states
    
    covered_goals = set()
    for p in selected:
        for atom in p.pattern:
            if atom in task_info.fluent_goal_atoms:
                covered_goals.add(atom)
    
    for goal in task_info.fluent_goal_atoms:
        if goal not in covered_goals:
            selected.append(Pattern(pattern=[goal]))
            total_states += 2  # 2^1 = 2 states
    
    if not selected and task_info.fluent_goal_atoms:
        for goal in task_info.fluent_goal_atoms:
            selected.append(Pattern(pattern=[goal]))
    
    return selected
\end{lstlisting}

\captionof{lstlisting}{Miconic generator code.}
\begin{lstlisting}[
    language=python,
    backgroundcolor=\color{backcolour},
    keywordstyle=\color{violet},
    basicstyle=\ttfamily\footnotesize,
    breakatwhitespace=false,
    breaklines=true,
    keepspaces=true,
    showspaces=false,
    numbers=none,
    showstringspaces=false,
    xleftmargin=0pt,
    xrightmargin=0pt,
]

def generate_pattern_collection(task_info: TaskInformation) -> list[Pattern]:
    
    patterns = []
    existing_patterns = set()  # Track to avoid duplicates
    
    atom_index = {}
    for atom in task_info.all_fluent_atoms:
        key = (atom.predicate.name, atom.binding)
        atom_index[key] = atom
    
    passenger_destin = {}
    for atom in task_info.static_ground_atoms:
        if atom.predicate.name == 'destin':
            p, f = atom.binding[0], atom.binding[1]
            passenger_destin[p] = f
    
    passenger_origin = {}
    for atom in task_info.fluent_initial_state_atoms:
        if atom.predicate.name == 'origin':
            p, f = atom.binding[0], atom.binding[1]
            passenger_origin[p] = f
    
    passenger_to_atoms = {}
    for atom in task_info.all_fluent_atoms:
        if atom.predicate.name in ('boarded', 'served', 'origin'):
            passenger = atom.binding[0]
            if passenger not in passenger_to_atoms:
                passenger_to_atoms[passenger] = []
            passenger_to_atoms[passenger].append(atom)
    
    lift_atoms = [atom for atom in task_info.all_fluent_atoms if atom.predicate.name == 'lift-at']
    
    for p, f_orig in passenger_origin.items():
        if p not in passenger_destin:
            continue
            
        f_dest = passenger_destin[p]
        pattern_atoms = []
        
        origin_key = ('origin', (p, f_orig))
        if origin_key in atom_index:
            pattern_atoms.append(atom_index[origin_key])
        
        boarded_key = ('boarded', (p,))
        if boarded_key in atom_index:
            pattern_atoms.append(atom_index[boarded_key])
        
        served_key = ('served', (p,))
        if served_key in atom_index:
            pattern_atoms.append(atom_index[served_key])
        
        lift_orig_key = ('lift-at', (f_orig,))
        if lift_orig_key in atom_index:
            pattern_atoms.append(atom_index[lift_orig_key])
        
        if f_dest != f_orig:  # Avoid duplicates if origin == destin
            lift_dest_key = ('lift-at', (f_dest,))
            if lift_dest_key in atom_index:
                pattern_atoms.append(atom_index[lift_dest_key])
        
        if len(pattern_atoms) >= 2:
            frozen_pat = frozenset(pattern_atoms)
            if frozen_pat not in existing_patterns:
                patterns.append(Pattern(pattern=pattern_atoms))
                existing_patterns.add(frozen_pat)
        
        for atom in pattern_atoms:
            if atom.predicate.name in ('boarded', 'served'):
                singleton = [atom]
                frozen_single = frozenset(singleton)
                if frozen_single not in existing_patterns:
                    patterns.append(Pattern(pattern=singleton))
                    existing_patterns.add(frozen_single)
    
    relevant_floors = set(passenger_origin.values()) | set(passenger_destin.values())
    relevant_lift_atoms = [atom for atom in lift_atoms if atom.binding[0] in relevant_floors]
    
    if relevant_lift_atoms:
        unique_floors = set(atom.binding[0] for atom in relevant_lift_atoms)
        if len(unique_floors) <= 20:
            frozen_pat = frozenset(relevant_lift_atoms)
            if frozen_pat not in existing_patterns:
                patterns.append(Pattern(pattern=relevant_lift_atoms))
                existing_patterns.add(frozen_pat)
        else:
            for atom in relevant_lift_atoms:
                singleton = [atom]
                frozen_single = frozenset(singleton)
                if frozen_single not in existing_patterns:
                    patterns.append(Pattern(pattern=singleton))
                    existing_patterns.add(frozen_single)
    
    floor_to_passengers = {}
    for passenger in passenger_to_atoms:
        if passenger in passenger_origin:
            floor = passenger_origin[passenger]
            floor_to_passengers.setdefault(floor, []).append(passenger)
        if passenger in passenger_destin:
            floor = passenger_destin[passenger]
            floor_to_passengers.setdefault(floor, []).append(passenger)
    
    for floor, passengers in floor_to_passengers.items():
        if len(passengers) <= 1:
            continue
        pattern_atoms = []
        for p in passengers[:6]:  # 6 passengers * ~3 atoms + 1 lift = ~19 variables
            pattern_atoms.extend(passenger_to_atoms[p])
        for atom in lift_atoms:
            if atom.binding[0] == floor:
                pattern_atoms.append(atom)
                break
        unique_vars = set()
        for atom in pattern_atoms:
            unique_vars.update(atom.binding)
        if len(unique_vars) <= 20:
            frozen_pat = frozenset(pattern_atoms)
            if frozen_pat not in existing_patterns:
                patterns.append(Pattern(pattern=pattern_atoms))
                existing_patterns.add(frozen_pat)
    
    if len(lift_atoms) > 1:
        floor_to_atom = {}
        floor_keys = []
        
        for atom in lift_atoms:
            floor_obj = atom.binding[0]
            try:
                floor_num = int(floor_obj.name[1:])
                floor_to_atom[floor_num] = atom
                floor_keys.append(floor_num)
            except (ValueError, IndexError):
                floor_to_atom[floor_obj.name] = atom
                floor_keys.append(floor_obj.name)
        
        try:
            sorted_floors = sorted(floor_keys)
        except TypeError:
            sorted_floors = sorted(floor_keys, key=str)
        
        for i in range(len(sorted_floors) - 1):
            f_curr, f_next = sorted_floors[i], sorted_floors[i + 1]
            atom_curr = floor_to_atom[f_curr]
            atom_next = floor_to_atom[f_next]
            
            adj_pattern = [atom_curr, atom_next]
            frozen_pat = frozenset(adj_pattern)
            if frozen_pat not in existing_patterns:
                patterns.append(Pattern(pattern=adj_pattern))
                existing_patterns.add(frozen_pat)
    
    goal_singletons_added = set()
    for pattern in patterns:
        if len(pattern.pattern) == 1:
            goal_singletons_added.add(pattern.pattern[0])
    
    for goal_atom in task_info.fluent_goal_atoms:
        if goal_atom not in goal_singletons_added:
            patterns.append(Pattern(pattern=[goal_atom]))
    
    if not patterns:
        for goal_atom in task_info.fluent_goal_atoms:
            patterns.append(Pattern(pattern=[goal_atom]))
    
    return patterns
\end{lstlisting}

\captionof{lstlisting}{Rovers generator code.}
\begin{lstlisting}[
    language=python,
    backgroundcolor=\color{backcolour},
    keywordstyle=\color{violet},
    basicstyle=\ttfamily\footnotesize,
    breakatwhitespace=false,
    breaklines=true,
    keepspaces=true,
    showspaces=false,
    numbers=none,
    showstringspaces=false,
    xleftmargin=0pt,
    xrightmargin=0pt,
]

def generate_pattern_collection(task_info: TaskInformation) -> list[Pattern]:
    patterns = []
    estimated_states = 0
    MAX_TOTAL_STATES = 500_000_000
    processed_sigs = set()
    
    def count_vars(atom_list):
        objs = set()
        for atom in atom_list:
            objs.update(atom.binding)
        return len(objs)
    
    def add_pattern(atom_list):
        nonlocal estimated_states
        if not atom_list:
            return
        
        n_vars = count_vars(atom_list)
        if n_vars > 20:
            return
        
        states = 2 ** n_vars
        if estimated_states + states > MAX_TOTAL_STATES:
            return
            
        key = frozenset((a.predicate.name, a.binding) for a in atom_list)
        if key in processed_sigs:
            return
        processed_sigs.add(key)
        
        patterns.append(Pattern(pattern=list(atom_list)))
        estimated_states += states
    
    fluent_atoms = list(task_info.all_fluent_atoms)
    
    atoms_by_pred = {}
    for atom in fluent_atoms:
        pred_name = atom.predicate.name
        if pred_name not in atoms_by_pred:
            atoms_by_pred[pred_name] = []
        atoms_by_pred[pred_name].append(atom)
    
    store_to_rover = {}
    camera_to_rover = {}
    obj_visible_from = {}  # objective -> set of waypoints where visible
    lander_location = None
    visible_waypoints = set()  # waypoints visible from lander
    
    for atom in task_info.static_ground_atoms:
        if atom.predicate.name == 'store_of':
            store_to_rover[atom.binding[0]] = atom.binding[1]
        elif atom.predicate.name == 'on_board':
            camera_to_rover[atom.binding[0]] = atom.binding[1]
        elif atom.predicate.name == 'visible_from':
            obj, wp = atom.binding[0], atom.binding[1]
            if obj not in obj_visible_from:
                obj_visible_from[obj] = set()
            obj_visible_from[obj].add(wp)
        elif atom.predicate.name == 'at_lander':
            lander_location = atom.binding[1]
    
    if lander_location:
        for atom in task_info.static_ground_atoms:
            if atom.predicate.name == 'visible':
                wp1, wp2 = atom.binding[0], atom.binding[1]
                if wp1 == lander_location:
                    visible_waypoints.add(wp2)
                if wp2 == lander_location:
                    visible_waypoints.add(wp1)
    
    rover_capabilities = {}
    for atom in task_info.static_ground_atoms:
        if atom.predicate.name == 'equipped_for_soil_analysis':
            rover = atom.binding[0]
            if rover not in rover_capabilities:
                rover_capabilities[rover] = set()
            rover_capabilities[rover].add('soil')
        elif atom.predicate.name == 'equipped_for_rock_analysis':
            rover = atom.binding[0]
            if rover not in rover_capabilities:
                rover_capabilities[rover] = set()
            rover_capabilities[rover].add('rock')
        elif atom.predicate.name == 'equipped_for_imaging':
            rover = atom.binding[0]
            if rover not in rover_capabilities:
                rover_capabilities[rover] = set()
            rover_capabilities[rover].add('imaging')
    
    image_goals_by_obj = {}
    soil_goals = []
    rock_goals = []
    
    for goal in task_info.fluent_goal_atoms:
        if goal.predicate.name == 'communicated_image_data':
            obj, mode = goal.binding[0], goal.binding[1]
            if obj not in image_goals_by_obj:
                image_goals_by_obj[obj] = []
            image_goals_by_obj[obj].append((goal, mode))
        elif goal.predicate.name == 'communicated_soil_data':
            soil_goals.append((goal, goal.binding[0]))
        elif goal.predicate.name == 'communicated_rock_data':
            rock_goals.append((goal, goal.binding[0]))
    
    for obj, goal_modes in image_goals_by_obj.items():
        have_images = []
        relevant_rovers = set()
        modes = []
        for goal, mode in goal_modes:
            modes.append(mode)
            for have_img in atoms_by_pred.get('have_image', []):
                if have_img.binding[1] == obj and have_img.binding[2] == mode:
                    have_images.append(have_img)
                    relevant_rovers.add(have_img.binding[0])
        
        for rover in relevant_rovers:
            if 'imaging' not in rover_capabilities.get(rover, set()):
                continue
                
            pattern_atoms = [gm[0] for gm in goal_modes]
            for have_img in have_images:
                if have_img.binding[0] == rover and have_img not in pattern_atoms:
                    pattern_atoms.append(have_img)
            
            relevant_cameras = set()
            for static_atom in task_info.static_ground_atoms:
                if static_atom.predicate.name == 'supports':
                    cam, cam_mode = static_atom.binding[0], static_atom.binding[1]
                    if cam_mode in modes and camera_to_rover.get(cam) == rover:
                        relevant_cameras.add(cam)
            
            for cal in atoms_by_pred.get('calibrated', []):
                if cal.binding[1] == rover and cal.binding[0] in relevant_cameras:
                    pattern_atoms.append(cal)
            
            relevant_wps = obj_visible_from.get(obj, set())
            priority_wps = relevant_wps & visible_waypoints
            secondary_wps = (relevant_wps | visible_waypoints) - priority_wps
            
            for at in atoms_by_pred.get('at', []):
                if at.binding[0] == rover and at.binding[1] in priority_wps:
                    pattern_atoms.append(at)
            for at in atoms_by_pred.get('at', []):
                if at.binding[0] == rover and at.binding[1] in secondary_wps and at not in pattern_atoms:
                    pattern_atoms.append(at)
            
            add_pattern(pattern_atoms)
            
            for at in atoms_by_pred.get('at', []):
                if at.binding[0] == rover and at.binding[1] in visible_waypoints:
                    comm_pattern = [gm[0] for gm in goal_modes] + [at]
                    for have_img in have_images:
                        if have_img.binding[0] == rover:
                            comm_pattern.append(have_img)
                    add_pattern(comm_pattern)
                    break
    
    for goal, waypoint in soil_goals:
        pattern_atoms = [goal]
        for have_soil in atoms_by_pred.get('have_soil_analysis', []):
            if have_soil.binding[1] == waypoint:
                rover = have_soil.binding[0]
                if 'soil' not in rover_capabilities.get(rover, set()):
                    continue
                pattern_atoms.append(have_soil)
                for at in atoms_by_pred.get('at', []):
                    if at.binding[0] == rover and at.binding[1] == waypoint:
                        pattern_atoms.append(at)
                for at in atoms_by_pred.get('at', []):
                    if at.binding[0] == rover and at.binding[1] in visible_waypoints and at not in pattern_atoms:
                        pattern_atoms.append(at)
                for empty_atom in atoms_by_pred.get('empty', []):
                    store = empty_atom.binding[0]
                    if store_to_rover.get(store) == rover:
                        pattern_atoms.append(empty_atom)
                for full_atom in atoms_by_pred.get('full', []):
                    store = full_atom.binding[0]
                    if store_to_rover.get(store) == rover:
                        pattern_atoms.append(full_atom)
        
        add_pattern(pattern_atoms)
        
        for rover in rover_capabilities:
            if 'soil' not in rover_capabilities[rover]:
                continue
            for at in atoms_by_pred.get('at', []):
                if at.binding[0] == rover and at.binding[1] in visible_waypoints:
                    comm_pat = [goal]
                    for have_soil in atoms_by_pred.get('have_soil_analysis', []):
                        if have_soil.binding[0] == rover and have_soil.binding[1] == waypoint:
                            comm_pat.append(have_soil)
                            break
                    if len(comm_pat) < 20:
                        comm_pat.append(at)
                    add_pattern(comm_pat)
                    break
    
    for goal, waypoint in rock_goals:
        pattern_atoms = [goal]
        for have_rock in atoms_by_pred.get('have_rock_analysis', []):
            if have_rock.binding[1] == waypoint:
                rover = have_rock.binding[0]
                if 'rock' not in rover_capabilities.get(rover, set()):
                    continue
                pattern_atoms.append(have_rock)
                for at in atoms_by_pred.get('at', []):
                    if at.binding[0] == rover and at.binding[1] == waypoint:
                        pattern_atoms.append(at)
                for at in atoms_by_pred.get('at', []):
                    if at.binding[0] == rover and at.binding[1] in visible_waypoints and at not in pattern_atoms:
                        pattern_atoms.append(at)
                for empty_atom in atoms_by_pred.get('empty', []):
                    store = empty_atom.binding[0]
                    if store_to_rover.get(store) == rover:
                        pattern_atoms.append(empty_atom)
                for full_atom in atoms_by_pred.get('full', []):
                    store = full_atom.binding[0]
                    if store_to_rover.get(store) == rover:
                        pattern_atoms.append(full_atom)
        
        add_pattern(pattern_atoms)
        
        for rover in rover_capabilities:
            if 'rock' not in rover_capabilities[rover]:
                continue
            for at in atoms_by_pred.get('at', []):
                if at.binding[0] == rover and at.binding[1] in visible_waypoints:
                    comm_pat = [goal]
                    for have_rock in atoms_by_pred.get('have_rock_analysis', []):
                        if have_rock.binding[0] == rover and have_rock.binding[1] == waypoint:
                            comm_pat.append(have_rock)
                            break
                    if len(comm_pat) < 20:
                        comm_pat.append(at)
                    add_pattern(comm_pat)
                    break
    
    rover_locs = {}
    for atom in atoms_by_pred.get('at', []):
        rover = atom.binding[0]
        if rover not in rover_locs:
            rover_locs[rover] = []
        rover_locs[rover].append(atom)
    
    for rover, locs in rover_locs.items():
        if 1 < len(locs):
            add_pattern(locs)
    
    rover_stores = {}
    for atom in atoms_by_pred.get('empty', []):
        store = atom.binding[0]
        if store in store_to_rover:
            rover = store_to_rover[store]
            if rover not in rover_stores:
                rover_stores[rover] = []
            rover_stores[rover].append(atom)
    
    for atom in atoms_by_pred.get('full', []):
        store = atom.binding[0]
        if store in store_to_rover:
            rover = store_to_rover[store]
            if rover not in rover_stores:
                rover_stores[rover] = []
            rover_stores[rover].append(atom)
    
    for rover, store_atoms in rover_stores.items():
        if 1 < len(store_atoms):
            add_pattern(store_atoms)
            
            for atom in store_atoms:
                if atom.predicate.name == 'full':
                    store = atom.binding[0]
                    r = store_to_rover.get(store)
                    if r:
                        for have_soil in atoms_by_pred.get('have_soil_analysis', []):
                            if have_soil.binding[0] == r:
                                add_pattern([atom, have_soil])
                        for have_rock in atoms_by_pred.get('have_rock_analysis', []):
                            if have_rock.binding[0] == r:
                                add_pattern([atom, have_rock])
    
    rover_cals = {}
    for atom in atoms_by_pred.get('calibrated', []):
        rover = atom.binding[1]
        if rover not in rover_cals:
            rover_cals[rover] = []
        rover_cals[rover].append(atom)
    
    for rover, cals in rover_cals.items():
        if 1 < len(cals):
            add_pattern(cals)
            
            for cal in cals:
                cam = cal.binding[0]
                for static_atom in task_info.static_ground_atoms:
                    if static_atom.predicate.name == 'calibration_target' and static_atom.binding[0] == cam:
                        target_obj = static_atom.binding[1]
                        for vis_atom in task_info.static_ground_atoms:
                            if vis_atom.predicate.name == 'visible_from' and vis_atom.binding[0] == target_obj:
                                cal_wp = vis_atom.binding[1]
                                for at in atoms_by_pred.get('at', []):
                                    if at.binding[0] == rover and at.binding[1] == cal_wp:
                                        add_pattern([cal, at])
                                        break
                        break
    
    covered_goals = set()
    for p in patterns:
        for atom in p.pattern:
            if atom in task_info.fluent_goal_atoms:
                covered_goals.add(atom)
    
    for goal in task_info.fluent_goal_atoms:
        if goal not in covered_goals:
            add_pattern([goal])
    
    if not patterns and task_info.fluent_initial_state_atoms:
        add_pattern([task_info.fluent_initial_state_atoms[0]])
    
    return patterns
\end{lstlisting}

\captionof{lstlisting}{Satellite generator code.}
\begin{lstlisting}[
    language=python,
    backgroundcolor=\color{backcolour},
    keywordstyle=\color{violet},
    basicstyle=\ttfamily\footnotesize,
    breakatwhitespace=false,
    breaklines=true,
    keepspaces=true,
    showspaces=false,
    numbers=none,
    showstringspaces=false,
    xleftmargin=0pt,
    xrightmargin=0pt,
]

def generate_pattern_collection(task_info: TaskInformation) -> list[Pattern]:
    patterns = []
    
    supports_map = {}  # mode -> list of instruments
    on_board_map = {}  # instrument -> satellite
    calib_target_map = {}  # instrument -> calibration direction
    
    for atom in task_info.static_ground_atoms:
        if atom.predicate.name == "supports":
            inst, mode = atom.binding[0], atom.binding[1]
            if mode not in supports_map:
                supports_map[mode] = []
            supports_map[mode].append(inst)
        elif atom.predicate.name == "on_board":
            inst, sat = atom.binding[0], atom.binding[1]
            on_board_map[inst] = sat
        elif atom.predicate.name == "calibration_target":
            inst, dir = atom.binding[0], atom.binding[1]
            calib_target_map[inst] = dir
    
    fluent_index = {}
    for atom in task_info.all_fluent_atoms:
        pred = atom.predicate.name
        if pred not in fluent_index:
            fluent_index[pred] = []
        fluent_index[pred].append(atom)
    
    def get_atom(pred_name, condition):
        for atom in fluent_index.get(pred_name, []):
            if condition(atom):
                return atom
        return None
    
    def count_variables(atom_list):
        vars = set()
        for atom in atom_list:
            for obj in atom.binding:
                vars.add(obj)
        return len(vars)
    
    def deduplicate_atoms(atom_list):
        seen = set()
        result = []
        for atom in atom_list:
            if atom not in seen:
                seen.add(atom)
                result.append(atom)
        return result
    
    pointing_by_sat = {}
    for atom in task_info.all_fluent_atoms:
        if atom.predicate.name == "pointing" and len(atom.binding) >= 2:
            sat = atom.binding[0]
            if sat not in pointing_by_sat:
                pointing_by_sat[sat] = []
            pointing_by_sat[sat].append(atom)
    
    for sat, point_atoms in pointing_by_sat.items():
        if len(point_atoms) > 1:
            deduped = deduplicate_atoms(point_atoms)
            if count_variables(deduped) <= 20:
                patterns.append(Pattern(pattern=deduped))
    
    processed_instruments = set()
    
    for goal in task_info.fluent_goal_atoms:
        if goal.predicate.name != "have_image":
            patterns.append(Pattern(pattern=[goal]))
            continue
        
        direction, mode = goal.binding[0], goal.binding[1]
        
        candidate_instruments = supports_map.get(mode, [])[:3]
        
        for inst in candidate_instruments:
            sat = on_board_map.get(inst)
            if not sat:
                continue
            
            pattern_atoms = [goal]
            
            calib = get_atom("calibrated", lambda a: a.binding[0] == inst)
            if calib:
                pattern_atoms.append(calib)
            
            power = get_atom("power_on", lambda a: a.binding[0] == inst)
            if power:
                pattern_atoms.append(power)
            
            point = get_atom("pointing", lambda a: a.binding[0] == sat and a.binding[1] == direction)
            if point:
                pattern_atoms.append(point)
            
            calib_dir = calib_target_map.get(inst)
            if calib_dir and calib_dir != direction:
                point_calib = get_atom("pointing", lambda a: a.binding[0] == sat and a.binding[1] == calib_dir)
                if point_calib and count_variables(pattern_atoms + [point_calib]) <= 20:
                    pattern_atoms.append(point_calib)
            
            pavail = get_atom("power_avail", lambda a: a.binding[0] == sat)
            if pavail and count_variables(pattern_atoms + [pavail]) <= 20:
                pattern_atoms.append(pavail)
            
            final_atoms = deduplicate_atoms(pattern_atoms)
            if len(final_atoms) > 1 and count_variables(final_atoms) <= 20:
                patterns.append(Pattern(pattern=final_atoms))
            
            if inst not in processed_instruments:
                processed_instruments.add(inst)
                calib_pattern = []
                
                if calib:
                    calib_pattern.append(calib)
                if power:
                    calib_pattern.append(power)
                if pavail:
                    calib_pattern.append(pavail)
                if calib_dir:
                    point_calib = get_atom("pointing", lambda a: a.binding[0] == sat and a.binding[1] == calib_dir)
                    if point_calib:
                        calib_pattern.append(point_calib)
                
                final_calib = deduplicate_atoms(calib_pattern)
                if len(final_calib) >= 2 and count_variables(final_calib) <= 20:
                    patterns.append(Pattern(pattern=final_calib))
    
    sat_to_instruments = {}
    for inst, sat in on_board_map.items():
        if sat not in sat_to_instruments:
            sat_to_instruments[sat] = []
        sat_to_instruments[sat].append(inst)
    
    for sat, insts in sat_to_instruments.items():
        power_pattern = []
        pavail = get_atom("power_avail", lambda a: a.binding[0] == sat)
        if pavail:
            power_pattern.append(pavail)
        for inst in insts:
            pon = get_atom("power_on", lambda a: a.binding[0] == inst)
            if pon:
                power_pattern.append(pon)
        if len(power_pattern) > 1 and count_variables(power_pattern) <= 20:
            patterns.append(Pattern(pattern=deduplicate_atoms(power_pattern)))
    
    goals_by_sat = {}
    for goal in task_info.fluent_goal_atoms:
        if goal.predicate.name == "have_image":
            direction, mode = goal.binding[0], goal.binding[1]
            for inst in supports_map.get(mode, []):
                sat = on_board_map.get(inst)
                if sat:
                    if sat not in goals_by_sat:
                        goals_by_sat[sat] = []
                    goals_by_sat[sat].append((goal, inst, direction))
    
    for sat, entries in goals_by_sat.items():
        if len(entries) > 1:
            combined = []
            pavail = get_atom("power_avail", lambda a: a.binding[0] == sat)
            if pavail:
                combined.append(pavail)
            
            for i, (goal, inst, direction) in enumerate(entries[:2]):
                if count_variables(combined + [goal]) > 20:
                    break
                combined.append(goal)
                
                calib = get_atom("calibrated", lambda a: a.binding[0] == inst)
                if calib and count_variables(combined + [calib]) <= 20:
                    combined.append(calib)
                
                power = get_atom("power_on", lambda a: a.binding[0] == inst)
                if power and count_variables(combined + [power]) <= 20:
                    combined.append(power)
                
                point = get_atom("pointing", lambda a: a.binding[0] == sat and a.binding[1] == direction)
                if point and count_variables(combined + [point]) <= 20:
                    combined.append(point)
            
            if len(combined) > 1:
                final_combined = deduplicate_atoms(combined)
                if count_variables(final_combined) <= 20 and len(final_combined) > 1:
                    patterns.append(Pattern(pattern=final_combined))
    
    covered_goals = set()
    for p in patterns:
        for atom in p.pattern:
            if atom in task_info.fluent_goal_atoms:
                covered_goals.add(atom)
    
    for goal in task_info.fluent_goal_atoms:
        if goal not in covered_goals:
            patterns.append(Pattern(pattern=[goal]))
    
    if not patterns and task_info.fluent_initial_state_atoms:
        patterns.append(Pattern(pattern=[task_info.fluent_initial_state_atoms[0]]))
    
    return patterns
\end{lstlisting}

\captionof{lstlisting}{Transport generator code.}
\begin{lstlisting}[
    language=python,
    backgroundcolor=\color{backcolour},
    keywordstyle=\color{violet},
    basicstyle=\ttfamily\footnotesize,
    breakatwhitespace=false,
    breaklines=true,
    keepspaces=true,
    showspaces=false,
    numbers=none,
    showstringspaces=false,
    xleftmargin=0pt,
    xrightmargin=0pt,
]

def generate_pattern_collection(task_info: TaskInformation) -> list[Pattern]:
    patterns = []
    
    packages = set()
    vehicles = set()
    
    for atom in task_info.fluent_goal_atoms:
        if atom.predicate.name == 'at':
            packages.add(atom.binding[0])
    
    for atom in task_info.all_fluent_atoms:
        if atom.predicate.name == 'in':
            packages.add(atom.binding[0])  # package
            vehicles.add(atom.binding[1])  # vehicle
        elif atom.predicate.name == 'capacity':
            vehicles.add(atom.binding[0])
    
    for pkg in packages:
        pkg_atoms = []
        
        for atom in task_info.fluent_initial_state_atoms:
            if atom.predicate.name == 'at' and atom.binding[0] == pkg:
                pkg_atoms.append(atom)
                break
        
        for atom in task_info.fluent_goal_atoms:
            if atom.predicate.name == 'at' and atom.binding[0] == pkg:
                pkg_atoms.append(atom)
                break
        
        for atom in task_info.all_fluent_atoms:
            if atom.predicate.name == 'in' and atom.binding[0] == pkg:
                pkg_atoms.append(atom)
        
        if pkg_atoms and len(pkg_atoms) <= 20:
            patterns.append(Pattern(pattern=pkg_atoms))
    
    for veh in vehicles:
        veh_atoms = []
        for atom in task_info.all_fluent_atoms:
            if atom.predicate.name == 'at' and atom.binding[0] == veh:
                veh_atoms.append(atom)
            elif atom.predicate.name == 'capacity' and atom.binding[0] == veh:
                veh_atoms.append(atom)
        
        if veh_atoms and len(veh_atoms) <= 20:
            patterns.append(Pattern(pattern=veh_atoms))
    
    for veh in vehicles:
        cargo_atoms = []
        for atom in task_info.all_fluent_atoms:
            if atom.predicate.name == 'in' and atom.binding[1] == veh:
                cargo_atoms.append(atom)
            elif atom.predicate.name == 'capacity' and atom.binding[0] == veh:
                cargo_atoms.append(atom)
        
        if cargo_atoms:
            cargo_atoms = cargo_atoms[:20]
            patterns.append(Pattern(pattern=cargo_atoms))
    
    critical_locations = set()
    
    for atom in task_info.fluent_initial_state_atoms:
        if atom.predicate.name == 'at' and atom.binding[0] in packages:
            critical_locations.add(atom.binding[1])
    
    for atom in task_info.fluent_goal_atoms:
        if atom.predicate.name == 'at' and atom.binding[0] in packages:
            critical_locations.add(atom.binding[1])
    
    for loc in critical_locations:
        loc_atoms = [atom for atom in task_info.all_fluent_atoms 
                     if atom.predicate.name == 'at' and atom.binding[1] == loc]
        
        if len(loc_atoms) > 20:
            priority_atoms = []
            for atom in loc_atoms:
                if atom in task_info.fluent_goal_atoms or atom in task_info.fluent_initial_state_atoms:
                    priority_atoms.append(atom)
            loc_atoms = priority_atoms[:20]
        
        if loc_atoms and len(loc_atoms) <= 20:
            patterns.append(Pattern(pattern=loc_atoms))
    
    covered_atoms = set()
    for pat in patterns:
        for atom in pat.pattern:
            covered_atoms.add(atom)
    
    for goal_atom in task_info.fluent_goal_atoms:
        if goal_atom not in covered_atoms:
            patterns.append(Pattern(pattern=[goal_atom]))
    
    if not patterns and task_info.fluent_initial_state_atoms:
        patterns.append(Pattern(pattern=[task_info.fluent_initial_state_atoms[0]]))
    
    return patterns
\end{lstlisting}

\end{document}